%% file: main.tex
\documentclass{article}

\usepackage{microtype}
\usepackage{graphicx}
\usepackage{subfigure}
\usepackage{booktabs} %
\usepackage{multirow}

\usepackage{hyperref}

\usepackage{amsmath}
\usepackage{amssymb}
\usepackage{cleveref}
\usepackage[ruled,vlined]{algorithm2e}
\newtheorem{statement}{Statement}
\usepackage{dblfloatfix} 
\usepackage{listings}

\newcommand{\state}{\mathbf{s}}
\newcommand{\sspace}{\mathcal{S}}
\newcommand{\aspace}{\mathcal{A}}
\newcommand{\action}{\mathbf{a}}
\newcommand{\muvec}{\boldsymbol{\mu}}
\newcommand{\stdvec}{\boldsymbol{\sigma}}
\newcommand{\epsbold}{\boldsymbol{\epsilon}}

\usepackage[accepted]{icml2021}

\icmltitlerunning{Low-Precision Reinforcement Learning}

\begin{document}

\twocolumn[
\icmltitle{Low-Precision Reinforcement Learning: \\ 
Running Soft Actor-Critic in Half Precision
}

\icmlsetsymbol{equal}{*}

\begin{icmlauthorlist}
\icmlauthor{Johan Bjorck}{to}
\icmlauthor{Xiangyu Chen}{to}
\icmlauthor{Christopher De Sa}{to}
\icmlauthor{Carla P. Gomes}{to}
\icmlauthor{Kilian Q. Weinberger}{to}
\end{icmlauthorlist}

\icmlaffiliation{to}{Department of Computer Science, Cornell University, USA}

\icmlcorrespondingauthor{Johan Bjorck}{njb225@cornell.edu}

\icmlkeywords{Machine Learning, ICML}

\vskip 0.3in
]

\printAffiliationsAndNotice{} %

\begin{abstract}

\input{text/abstract}

\end{abstract}
\input{text/intro}
\input{text/background}

\input{text/proposed_method}

\input{text/experiments}

\input{text/related_concl}
\bibliographystyle{icml2021}
\bibliography{refs}
\clearpage
\appendix
\newpage
\input{text/appendix}

\end{document}

%% file: text/abstract.tex
Low-precision training has become a popular approach to reduce compute requirements, memory footprint, and energy consumption in supervised learning. In contrast, this promising approach has not yet enjoyed similarly widespread adoption within the reinforcement learning (RL) community, partly because RL agents can be notoriously hard to train even in full precision. In this paper we consider continuous control with the state-of-the-art SAC agent and demonstrate that a na\"ive adaptation of low-precision methods from supervised learning fails. We propose a set of six modifications, all straightforward to implement, that leaves the underlying agent and its hyperparameters unchanged but improves the numerical stability dramatically. The resulting modified SAC agent has lower memory and compute requirements while matching full-precision rewards, demonstrating that low-precision training can substantially accelerate state-of-the-art RL without parameter tuning.

%% file: text/intro.tex
\section{Introduction}

Reinforcement learning (RL) is a promising paradigm for constructing autonomous agents for transportation~\cite{kendall2019learning}, robotics~\cite{zhu2020ingredients},  scientific applications~\cite{bellemare2020autonomous} and beyond~\cite{silver2016mastering}. To enable such impressive feats, RL often requires large amounts of computation and samples from the underlying environments~\cite{silver2016mastering, jaderberg2019human, berner2019dota}. Consequently, many approaches for scaling RL have been proposed, including distributed~\cite{horgan2018distributed, wijmans2019dd} and asynchronous training \cite{mnih2016asynchronous}, GPU-accelerated environments~\cite{dalton2019accelerating} and many others~\cite{petrenko2020sample, liu2020high}. Within supervised learning, an emerging trend for accelerating deep learning is low-precision training, where one uses fewer than 32 bits to represent individual parameters, activations, and gradients~\cite{gupta2015deep}. This speeds up computations and decreases latency, memory footprint, and energy consumption~\cite{de2017understanding, reuther2019survey, gong2018highly}. Such benefits can accelerate RL research, allow longer deployment before recharging batteries~\cite{tomy2019battery}, lower latency in high-speed applications~\cite{liu2020deep}, and is especially relevant in RL from raw pixel observations which can be compute-intensive~\cite{zhan2020framework}. Despite such upsides, the RL community has not adopted low-precision training, in part as RL agents are notoriously hard to train even in full precision~\cite{henderson2017deep, engstrom2020implementation, jordan2020evaluating}.

The goal of this paper is to demonstrate the feasibility of RL in low precision without changing hyperparameters or the underlying RL algorithm. Specifically, we focus on continuous control environments~\cite{tassa2020dmcontrol} and consider the state-of-the-art soft actor-critic (SAC) algorithm~\cite{haarnoja2018soft}. We demonstrate that strategies developed for supervised learning, such as loss scaling and mixed-precision~\cite{micikevicius2017mixed}, fail in this context. To enable RL in low precision, we propose a set of six methods that improve the numerical stability without changing the underlying agent. Some of these are novel, for example, we propose to store the square root of the second moment in Adam to decrease the needed dynamic range. Others have more of an engineering flavor, e.g. reordering arithmetic operations, but are nonetheless crucially needed for performant agents.

We experimentally verify that with these methods it is possible to train SAC agents in low precision with comparable performance to full-precision agents, thus demonstrating the feasibility of low-precision RL. We further benchmark compute time and memory consumption and find dramatic improvements in both aspects. We perform experiments on RL from both environment state representations and pixel observations, and also perform ablation experiments and simulate various numerical formats with qtorch~\cite{zhang2019qpytorch}. Finally, we release our code to encourage future work on RL in low precision. Debugging numerical issues can be problematic as they do not necessarily result in crashes. We hope that the techniques and code we introduce can provide a starting point for other researchers interested in low-precision RL. 

%% file: text/background.tex
\section{Background}

\subsection{Low-Precision Deep Learning}

Within deep learning, model parameters are typically represented as 32-bit floating-point numbers (fp32)~\cite{kahan1996ieee}. One bit is used for the sign, the exponent uses 8 bits and the rest represent the significand. If we interpret these bits as integers $\text{s}, \text{e}, \text{m}$, the value they represent is %
$$
\text{value}(\text{s}, \text{e}, \text{m}) = (-1)^{\text{s}} \times 2^{\text{e} - 127} \times 1. \text{m}
$$
Arithmetic in low precision is precarious. With a finite number of bits, only a finite set of numbers can be represented. Calculations yielding numbers too large to be represented cause \textit{overflow} and result in $\infty$ (which has a specific bit-pattern). Conversely, numbers too small to be represented cause \textit{underflow} and are cast to 0. If $a$ is small and $b$ is large, in low precision the representable number closest to $a+b$ might be $b$. This can effectively undo addition. We refer to such issues as \textit{precision} issues. Finally, there is also a bit pattern for \textit{NaN} (abbreviation for "not a number") which can come from division by 0 or arithmetic with $\infty$.

\begin{figure}
\centering
\includegraphics[width=0.5\textwidth]{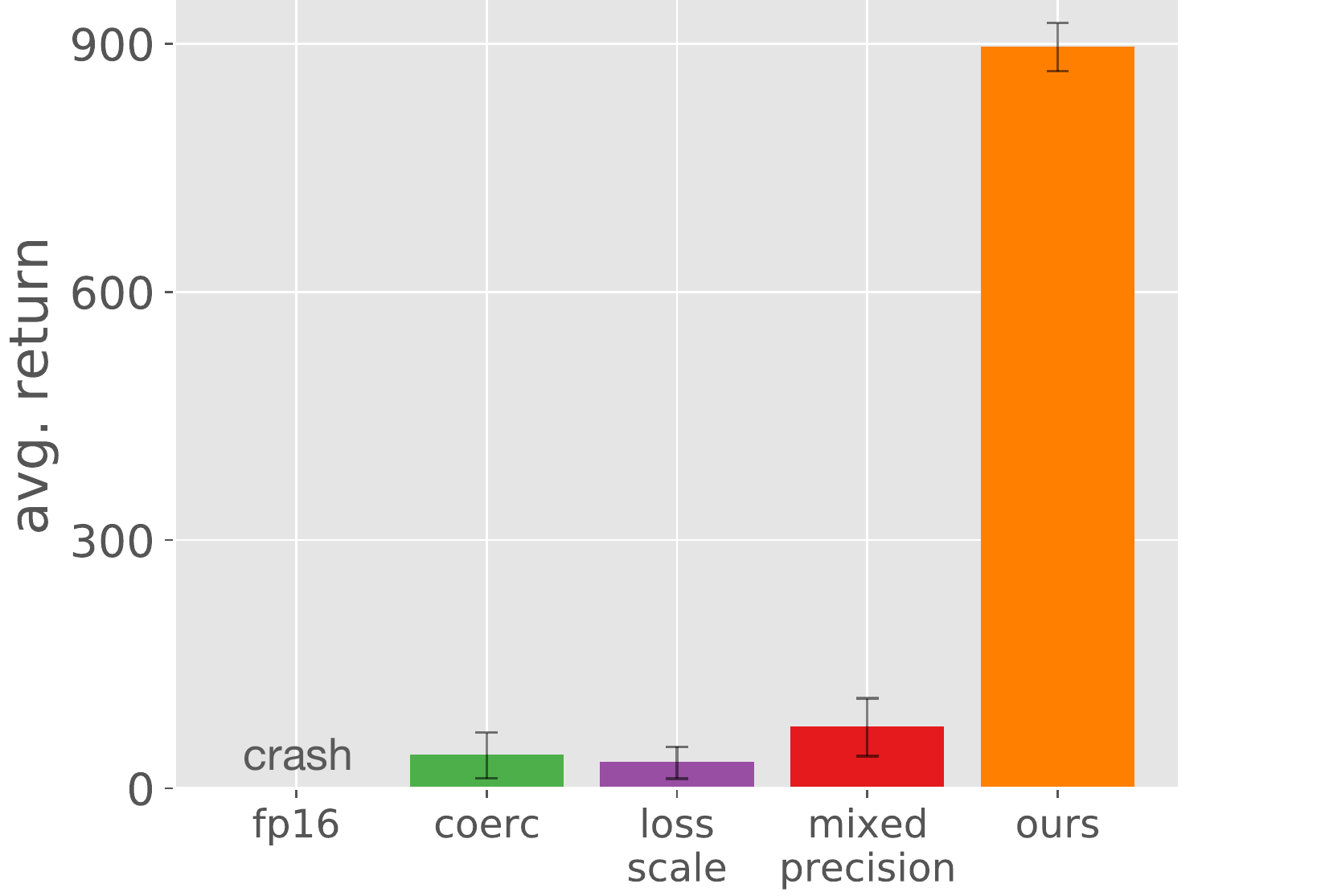}
\vspace{-1em}
\caption{Returns for the SAC agent, averaged across tasks in the planet-benchmarks, when using low-precision methods from supervised learning \cite{micikevicius2017mixed}. Naively adopting these methods in RL gives poor performance, suggesting that novel methods are needed. Fp16 always crashes, yielding no rewards.}
\vspace{-1em}
\label{fig:baseline}
\end{figure}

Low-precision deep learning relies on using fewer than 32 bits for individual parameters, activations, and gradients~\cite{gupta2015deep}, yielding faster computation, lower latency, less power consumption, and smaller memory footprint~\cite{de2017understanding, gong2018highly}. Nvidia GPUs~\cite{whitehead2011precision} and PyTorch~\cite{paszke2019pytorch} currently support 16-bit floating-point numbers (fp16 or half-precision), which uses 5 bits in the exponent~\cite{kahan1996ieee}. Other representations are actively being researched~\cite{kalamkar2019study, das2018mixed, song2017computation}. For low-precision training of neural networks, a few common tricks are used to improve numerical stability. One can scale the loss to prevent gradient underflow in the backward pass, or mix high and low precision~\cite{micikevicius2017mixed}, or coerce Nan values to 0 and $\infty$ to some large number. In \Cref{fig:baseline} we apply such baseline methods to our RL setting -- the SAC agent evaluated on the planet benchmark \cite{hafner2019learning} -- and observe that they do not reach competitive performance (see \Cref{sec:experiments} for details.) This suggests that novel strategies are needed to enable low-precision RL.

\subsection{Reinforcement Learning}

Continuous control tasks, e.g., robotic arm control, are often formulated as Markov decision processes (MDPs) defined by a tuple $(\sspace, \aspace, P, r)$~\cite{sutton2018reinforcement}. The state space $\sspace$ and action space $\aspace$ are both continuous and might be bounded. At each timestep $t$, the agent is in a state $\state_t\! \in\! \sspace$ and takes an action $\action \!\in\! \aspace$ to arrive at a new state $\state_{t+1} \in \sspace$. The transition between states given an action is random with transition probability $P : \sspace \times \sspace \times \aspace \rightarrow [0, \infty)$. Furthermore, at each timestep, the agent receives a reward $r_t$ as per the reward distribution $r$. The agent is typically trained to take actions that maximize expected discounted reward $\mathbb{E}  [\sum_i \gamma^i r_i]$ for some fixed $0\!<\!\gamma \!<\! 1$ which defines a horizon for the problem. The performance is typically measured by the cumulative rewards $\sum_i r_i$.

\subsection{Soft Actor-Critic (SAC)}

Soft actor-critic (SAC) is a popular RL algorithm for continuous control~\cite{haarnoja2018soft}, forming the basis for recent state-of-the-art algorithms such as CURL~\cite{laskin2020curl}, RAD~\cite{laskin2020reinforcement} and DRQ~\cite{kostrikov2020image}. For every state $\state$, SAC defines a continuous multivariate \textit{policy} distribution $\pi_\theta(\state)$ over actions, defined via a neural network with parameters $\theta$. 
The neural network outputs two vectors, a mean  $\muvec_\theta(\state)$ and a standard deviation  $\stdvec_\theta(\state)$ and the policy is defined as
\begin{equation}
\label{eq:sac_policy}
\pi_\theta(\state) = \tanh ( \muvec_\theta(\state) + \epsbold \odot \stdvec_\theta(\state)), \qquad \epsbold \sim \mathcal{N} (0, \mathbf{1}).
\end{equation}
Here, $\tanh$ is applied elementwise, yielding the action space $[-1,1]^n$ for some $n = {dim(\aspace)}$. An action $\action$ can then be sampled from $\pi_\theta(\state)$ simply by sampling $\epsbold$ from the standard normal distribution. 
To encourage \textit{exploration}, the entropy $H(\pi_\theta(\state_t ))$ of the action distribution (times a hyperparameter $\alpha$) is added to the reward at each state $\state_t$.  In practice, the entropy can be calculated by sampling an action $\action$ from $\pi_\theta$ and thereafter calculating $\log \pi_\theta( \action | \state) $. 

The agent also uses a neural network with parameters $\psi$, often referred to as the \textit{critic}, that outputs a q-value $Q_\psi(\state , \action)$ \cite{watkins1989learning} for each state $\state$ and action $\action$. Q-values roughly measure the expected rewards that can be obtained by taking action $\action$ in state $\state$. The critic network is trained to minimize the square loss between its output $Q_\psi(\action, \state_t)$ and the target $r_t + \gamma \mathbb{E} [ \hat{Q}(\action_{t+1}, \state_{t+1}) + \alpha H(\pi)]$. The value $\hat{Q}(\state, \action)$ is obtained from another network -- the \textit{target} network --  which shares architecture with the critic but whose weights $\hat{\psi}$ are the exponentially averaged weights of the critic network, i.e., we do not take gradients w.r.t $\hat{\psi}$ and for some $\beta$ we occasionally update 
$$
\hat{\psi} \leftarrow  \beta \hat{\psi} + (1-\beta) \psi.
$$
The policy $\pi_\theta$ is optimized to maximize the expected rewards. This can be done by sampling an action $\action$ from $\pi_\theta(\state_t)$ and thereafter taking derivatives w.r.t. $\action$ \textit{through} the q-values $Q_\psi(\state, \action)$. In practice, instead of treating $\alpha$ as a fixed hyperparameter, one can update $\alpha$ with a gradient descent type optimizer so that the average entropy of states matches some predefined value. SAC is typically optimized by the Adam optimizer \cite{kingma2014adam}. For further details, see~\citet{haarnoja2018soft}.

%% file: text/proposed_method.tex
\begin{table}[h]
\vspace{-1em}
\caption{Proposed modifications and the problems they solve.}
\label{tab:methods}
\begin{center}
    \begin{tabular}{ l | l | l }
    \hline
    \# & method & problem \\ \hline
    1 & hAdam & under/overflow \\ 
    2 & softplus-fix & overflow \\ 
    3 & normal-fix & underflow \\
    4 & Kahan-momentum & precision/underflow \\
    5 & compound loss scaling & underflow \\
    6 & Kahan-gradients & precision \\
    \hline
    \end{tabular}
    \vspace{-1em}
\end{center}
\end{table}

\section{Proposed Methods}

\label{sec:proposed_method}

To improve the numerical stability of SAC \cite{haarnoja2018soft} we propose a set of six modifications that do not change the underlying algorithm. These are listed in \Cref{tab:methods}. We point out that some of them are relatively straightforward engineering choices; nonetheless, such implementation level modifications are often important to achieve good performance in RL \cite{engstrom2020implementation}. For the sake of being self-contained, we present all modifications here. We verify that these components individually contribute to improved performance in~\Cref{fig:ablation}. However, not all these components are needed for all environments -- we have found that some environments are less sensitive to numerical issues; see \cref{sec:app_more_experiments} for details. Many of these methods are agnostic to the underlying RL method and we hope that they provide a sensible starting point for other RL agents.

\textbf{1. Storing the Hypotenuse in Adam.} The Adam optimizer stores the exponential moving average of the squared gradient. It is stored in a variable $v$ which is updated as $v_{t+1} \leftarrow \beta_2 v_t + (1-\beta_2) g_{t+1}^2$ for gradient $g_{t+1}$. Storing the full range of $v$ is infeasible in low precision; for example, if $g_{t+1} = 10^{-7}$ computing $g_{t+1}^2$ might cause underflow in 16-bit precision. We instead propose to store $w = \sqrt{v}$ which has a smaller dynamic range than $v$. To update $w$ we use the hypot function, which is defined as
$$
\text{hypot}(a,b) = \sqrt{a^2 + b^2}.
$$
In a na\"ive implementation, the intermediate results $a^2$ or $b^2$ might underflow. To avoid this, we can ,e.g., rewrite the expression above as follows for nonzero $a$ and $b$:
$$
\text{hypot}(a,b) = \max(a,b) \sqrt{1 + \bigg(\frac{\min(a,b)}{\max(a,b)} \bigg)^2}.
$$
This expression is more numerically stable when $a$ and $b$ can be represented, but $a^2$ and $b^2$ are rounded to $0$. To allow for $a=0, b=0$ one can add a numerical $\epsilon$ to the denominator. Once we have a numerically suitable hypot function we define the updates for $w$ as follows
$$
w_{t+1} \gets \text{hypot}( \sqrt{\beta_2} w_t, \sqrt{(1-\beta_2)} g_{t+1}).
$$
Under this update rule $w$ retains the semantics $w=\sqrt{v}$. We update the network parameters $\theta$ with $\theta \gets \theta  - \alpha \frac{m_t}{w_t + \epsilon}$ for some learning rate $\alpha$ and the $\epsilon$ used in Adam. In practice, this exchanges the square root computation to the more numerically stable hypot execution. Both have comparable complexity as they require finding a square root via iterative methods. In fact, hypot is a common numerical primitive (found in, e.g., $<$math.h$>$ in the C language).  We can pre-compute the values of $\sqrt{\beta_2}$ and $\sqrt{(1-\beta_2)}$ up-front.  Although this modification might cause a slight computational overhead, depending on the hypot implementation, the gradient computation dominates the computational cost. The resulting method, which we call \textbf{hAdam}, is described in~\Cref{alg:nadam}. 

\begin{algorithm}[H]
\SetAlgoLined
 $ m \gets 0$ \\
 $ w \gets 0$ \\
\For{$t = 0, 1, 2 ....$}{
$g \gets \nabla_\theta \ell $ \\
$m \gets \beta_1 m + (1-\beta_1) g $ \\
$w \gets \text{hypot}( \sqrt{\beta_2} w, \sqrt{(1-\beta_2)} g) $ \\
$ m \gets m / (1-\beta_1^t)$ // bias correction  \\
$ w \gets w / \sqrt{(1-\beta_2^t)}$ // bias correction  \\
$\theta \gets \theta - \alpha m / (w + \epsilon)$
}
\caption{hAdam.}
\label{alg:nadam}
\end{algorithm}

\textbf{2 \& 3. Numerical Issues in the Policy.} Na\"ively  handling the policy parametrization (\cref{eq:sac_policy}) of SAC leads to poor performance in low precision. 
Calculating the log-probability of actions, which is needed to compute the entropy, is problematic in low precision. Recall that actions $\action$ are drawn from $\tanh ( \mathbf{u})$ where $\mathbf{u} = \muvec_\theta + \epsbold \odot \stdvec_\theta$ and $\epsbold \sim \mathcal{N}(0,1)$. Here, $\muvec_\theta$ and $\stdvec_\theta$ are the outputs from the policy neural network. To calculate the log-probability, we need to account for the change of variables due to the tanh transformation:
\begin{align}
\log \pi ( \action | \state )  = \log P( \mathbf{u} | \state ) - \sum_i \log \big(1 - \tanh^2(u_i) \big) \nonumber\\ 
 = \log P ( \mathbf{u} | \state )  - \sum_i 2 \big[ \log(2) - u - \log \big( 1 + \exp(-2 u_i) \big) \big].\nonumber
\end{align}
The first expression is numerically unsuitable as $\tanh^2(u_i)$ can be rounded to 1 with insufficient precision. 
The latter expression is known to be more numerically stable~\cite{kostrikov2020image}, but can still overflow in the backward pass when $\exp(-2 u_i)$ is large. This happens, e.g., in PyTorch \cite{paszke2019pytorch}. 
To prevent such overflow, we exchange $\log \big( 1 + \exp(-2 u_i) \big)$ by a linear function (which has a stable backward pass) for sufficiently large negative $u_i$, i.e when $u < K$ for a $K$ that is chosen depending on the dynamic range of the number representation. The resulting function is shown in \cref{eq:softplus_fixed}. This modification only carries a minimal computational overhead and we call it \textbf{softplus-fix}.
\begin{equation}
\label{eq:softplus_fixed}
\text{softplus}'(u_i) = \begin{cases}
-  2 u_i  &\text{if $u < K$}  \\
\log \big( 1 + \exp(- 2 u_i) \big) &\text{ else}
\end{cases}
\end{equation}
Secondly, we need to calculate $\log P( \mathbf{u} | \state)$ which is just the log-probability of a normal variable. The numerical stability of this operation is important, and, e.g., the PyTorch implementation of the normal distribution has log-probabilities calculated as $\log x \sim \frac{(x-\mu)^2}{\sigma^2}$ (we omit normalization constants here). If $x, \mu, \sigma$ are all very small, it is possible that, e.g., $\sigma^2$ is rounded to zero in 16-bit precision despite the correct ratio being around 1. To improve the stability, we instead calculate the log-probability as $\log x \sim \big( \frac{x-\mu}{\sigma} \big)^2$ (modulo normalization constant). This turns out to solve such underflow issues for the SAC implementation of \citet{pytorch_sac} and incurs no computational or memory cost. We call this modification the \textbf{normal-fix}.

\textbf{4. Stability of Soft-Updates.} Recall that the target network is an exponential moving average of the critic network:
$$
\hat{\psi}\gets  \beta \hat{\psi}+ (1-\beta) \psi.
$$
For $\beta$ close to one, $(1-\beta) \psi$ can be very small which can cause numerical issues. To compute the momentum update of the target network, we propose to use Kahan summation \cite{kahan1965further}, a general method for calculating sums of low-precision numbers. When summing a long sequence of numbers in low precision, the numerical error can grow with the length of the sequence. Kahan summation adds a compensation variable that is updated as each term of the sum is added, which improves numerical stability. The method is given in \cref{alg:kahan}. We rewrite the momentum update as adding $(1-\beta )(\psi - \hat{\psi})$ to $\hat{\psi}$ and then use Kahan summation to add this term to $\hat{\psi}$. To prevent underflow of $(1-\beta )(\psi - \hat{\psi})$ we can scale it by some constant $ C >1$ and add them to a copy of $\hat{\psi}$ where all weights are scaled by $C$. This strategy has a relatively small computational cost but larger memory costs, however, this is offset by storing the model in lower precision. Additionally, we still reap the benefits of computing gradients and activations in lower-precision, which often is the main source of memory consumption. We call this method \textbf{Kahan-momentum}.

\begin{algorithm}[H]
\SetAlgoLined
 $ sum \gets 0$ \\
 $ c \gets 0$ \\
\For{val in values}{
$y \gets val - c $ \\
$ t \gets sum + y$ \\
$  c \gets (t - sum) - y$ \\
$ sum \gets t$
}
\caption{Kahan summation \cite{kahan1965further}.}
\label{alg:kahan}
\end{algorithm}

\begin{figure*}[b!]
\centering
\vspace{-1em}
\includegraphics[width=0.95\textwidth]{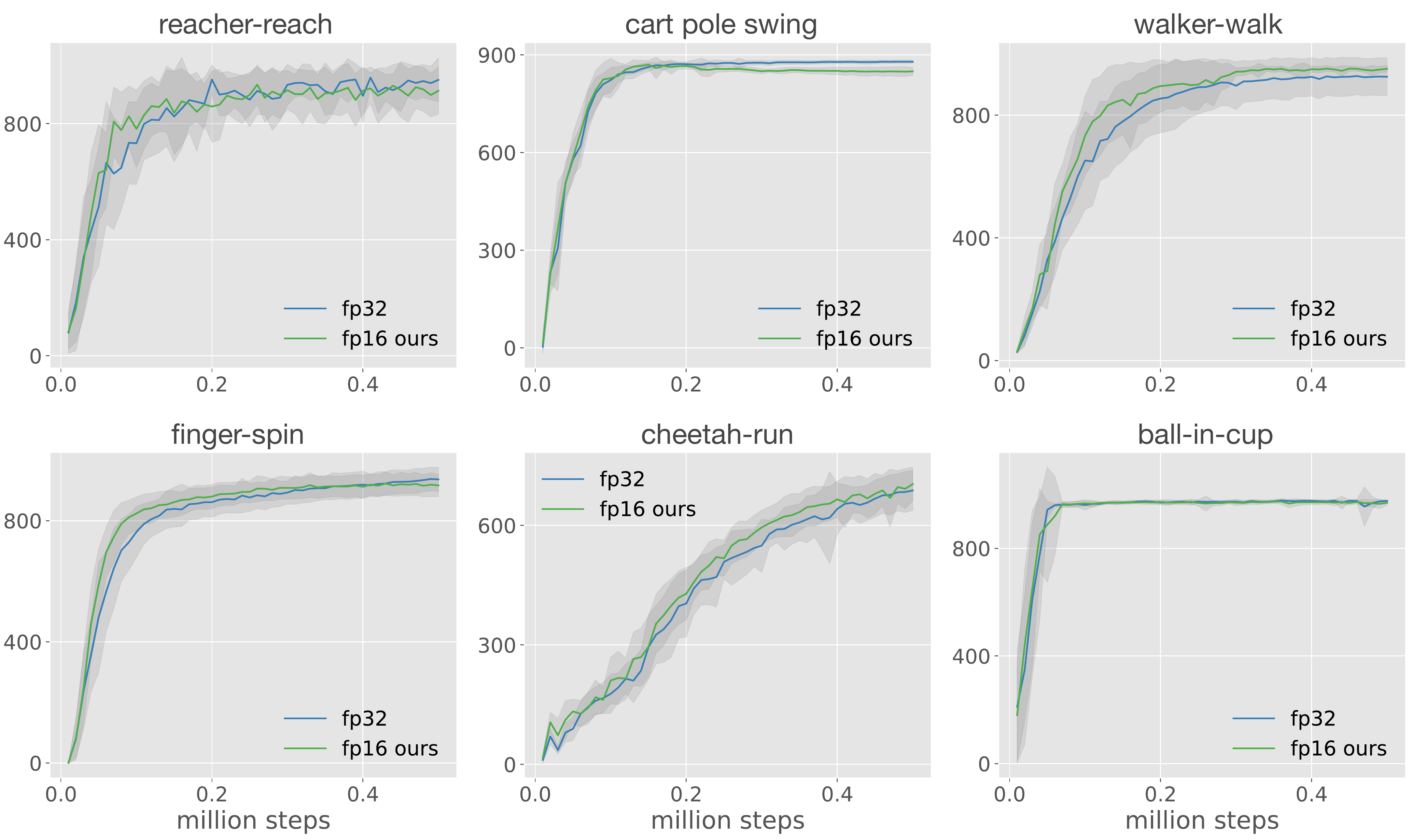}
\caption{Learning curves for SAC \cite{haarnoja2018soft} on the planet-benchmark for standard fp32 training and training in fp16 with our proposed methods. Naive fp16 methods fail as per \Cref{fig:baseline}. Performance is very close for all environments, demonstrating that low-precision RL is possible without changing the underlying RL algorithm or its hyperparameters.}
\vspace{-1em}
\label{fig:planet_bench}
\end{figure*}

\textbf{5. Compound Loss Scaling.} Adam stores the exponential moving averages of the gradients and the squared gradients, updating them as, e.g., $m \gets \beta_1 m + (1-\beta_1)g$. For small gradients $g$ and $\beta_1$ close to 1, the term $(1-\beta_1)g$ might be rounded to zero. Fortunately, the Adam update is invariant if we rescale the numerator and the denominator simultaneously. Thus, we can choose some fixed large $\gamma$ and update the Adam buffers as follows
$$
m \gets \beta_1 m + (1-\beta_1) \gamma g,
$$
$$
w \gets \text{hypot}( \sqrt{\beta} w, \sqrt{(1-\beta)} g \gamma ).
$$
This modifications ensure that $(1-\beta_1) \gamma g$ does not underflow. To retain the semantics of Adam, we update the parameters as $\theta \gets \theta - \alpha \frac{m}{w + \gamma \epsilon}$. For implementing this strategy we scale the loss by some $\gamma$ -- this scales the gradients and thus the contents of $m$ and $w$. Due to its similarity with loss scaling, we call this method \textbf{compound loss scaling}. Compared to loss scaling we do not need to "unscale" the gradients. As is standard in loss scaling, we change $\gamma$ dynamically following the strategy of PyTorch amp \cite{amp}; see \cref{sec:hyperparams} for details. This results in a small extra computational cost as we need to check that the gradients do not overflow when dynamically updating $\gamma$.

\textbf{6. Kahan Summation of Gradients.} Lastly, we perform Kahan summation on the gradient updates for the weights $\psi$ of the critic network and for the parameter $\alpha$. This trick turns out not to be needed for the actor-network. Formally, for a gradient update $\theta \gets \theta + \Delta \theta$, we apply \cref{alg:kahan}. Again, this method comes with a small computational overhead but needs a memory footprint comparable to that of the actual model. We already save some memory from storing the parameters in low precision, and the main source of memory consumption is often computing the gradients, which is not affected. We call this method \textbf{Kahan-gradients}.

\subsection{Guarantees}

In practice, our modifications improve numerical stability. However, one might be worried that the underlying RL algorithm is changed. As the methods rely on algebraically rewriting operations, the underlying algorithm will be the same -- at least in infinite numerical precision where arithmetic is commutative and associative. We formalize this intuition as follows; see \Cref{sec:appn_proof} for a proof.
\begin{statement}
\label{th:main}
In infinite precision, training SAC with the modification (normal-fix, softplus-fix,  hAdam,  Kahan-momentum,  compound loss scaling, and Kahan-gradients) is equivalent to training without them.
\end{statement}

%% file: text/experiments.tex
\section{Experiments}

\label{sec:experiments}

\subsection{Setup}

To verify the efficacy of our proposed methods we now perform experiments on the state-of-the-art agent SAC~\cite{haarnoja2018soft}. We here focus on three specific questions about our method i) does it match rewards from fp32 training? ii) does it improve compute/memory consumption? iii) are all proposed components needed? We start from the publicly accessible SAC implementation of \citet{pytorch_sac} which is used in recent state-of-the-art algorithms such as CURL~\cite{laskin2020curl}, RAD \cite{laskin2020reinforcement} and DRQ \cite{kostrikov2020image}. For environments, we consider the planet benchmark, popularized by~\cite{hafner2019learning} and used in e.g. \citet{kostrikov2020image, laskin2020curl, laskin2020reinforcement}. It consists of six continuous control tasks from the deep mind control suite \cite{tassa2020dmcontrol}: finger spin, cartpole swingup, reacher easy, cheetah run, walker walk, and ball in cup catch. We do not tune hyperparameters, but instead use hyperparameters from~\citet{pytorch_sac} (listed in~\Cref{sec:hyperparams}). We measure performance after 500,000 environment steps as per \cite{kostrikov2020image, laskin2020curl, lee2019stochastic} with individual runs scored by the average return over 10 episodes. We use 15 seeds for the main experiments (\Cref{fig:planet_bench} and \Cref{fig:from_pixels}) and 5 seeds for all other experiments. Runs using SAC na\"ively ported to fp16 can crash when taking actions that are non-finite, these are scored as 0. Figures show means and standard deviations (stds) across runs; when aggregating stds across tasks, we calculate the stds of individual tasks and then plot the average std. We primarily perform experiments with fp16 as this format is natively supported in PyTorch and on Nvidia V100 GPUs. Fp16 has both a smaller dynamic range and lower precision compared to fp32. We simulate other numerical formats in \cref{sec:qtorch}. 

\subsection{Comparison to Fp32}

We first compare the performance of the SAC agent trained in fp32 and fp16, the latter using all our methods from \cref{sec:proposed_method}. Learning curves for different tasks are given in~\autoref{fig:planet_bench}. We see that half-precision training performs very close to full-precision training, validating that our methods essentially match fp32 performance despite training in half-precision.

\subsection{Comparison to Baselines}

We now compare our strategy to the following standard low-precision methods from supervised learning:
\begin{itemize}
    \item Numeric coercion, abbbreviated as coerc. NaNs are coerced into zero and $\pm \infty$ to the largest/smallest fp16 value. 
    \item Loss scaling \cite{micikevicius2017mixed}, abbreviated as loss scale. We perform loss scaling with dynamic scale, using the same parameters as for our compound scaling. In \cref{sec:app_more_experiments} we use the amp \cite{amp} default parameters which gives similar results.
    \item Mixed precision and loss scaling \cite{micikevicius2017mixed}, abbreviated as mixed precision. In addition to loss scaling, we use mixed precision. Gradients, Adam buffers, and parameters are stored in fp32. Parameters are cast to half-precision in the forward pass, and the backward pass is calculated in half-precision.
\end{itemize}
In \Cref{fig:baseline} we show the performance of these baselines from supervised learning. These methods perform poorly in low-precision RL, demonstrating that additional strategies are necessary in this context.

\begin{figure}
\centering
\includegraphics[width=0.45\textwidth]{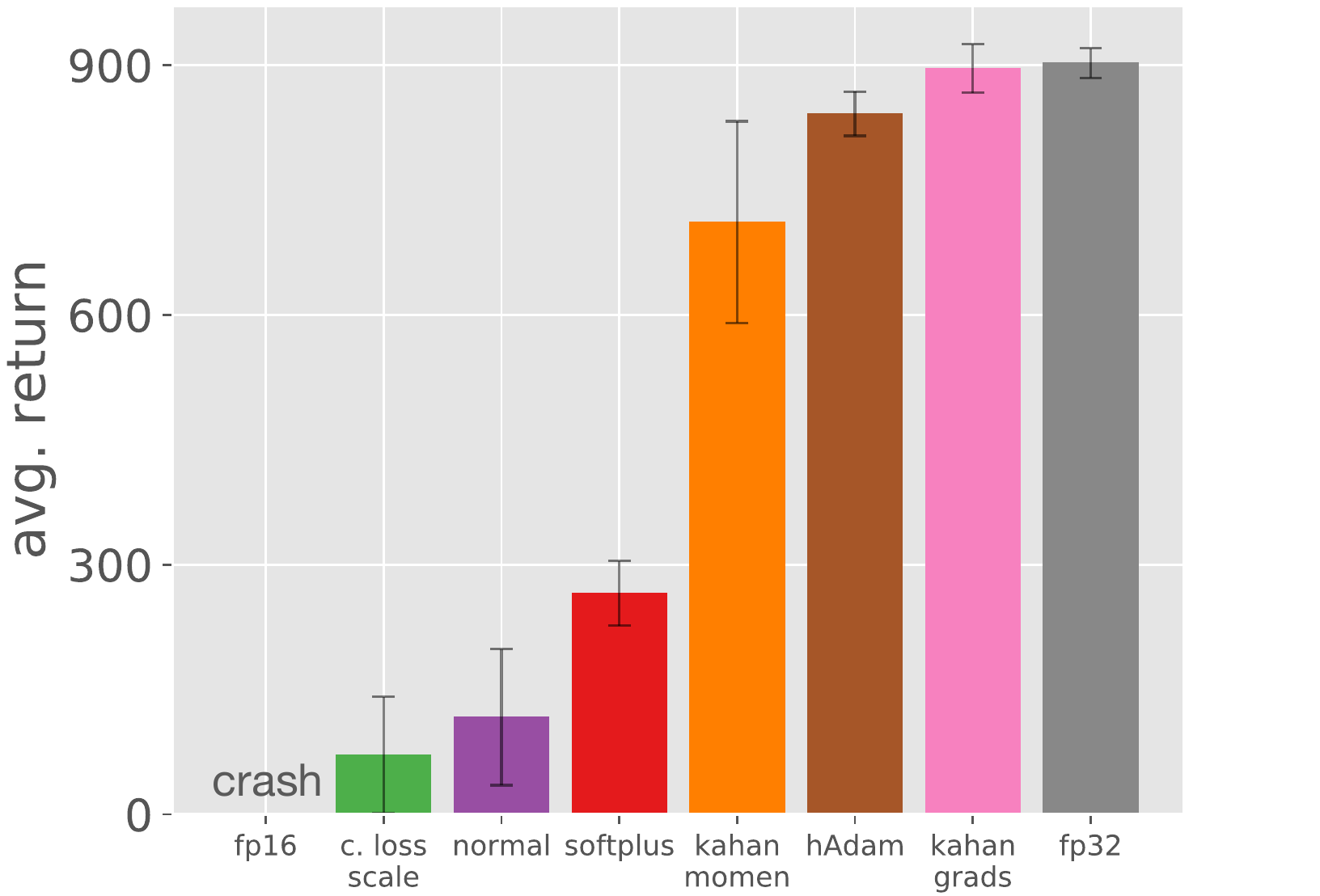}
\vspace{-0.5\baselineskip}
\caption{Ablation experiment where we add our components one-by-one until the complete agent is reached. Performance is averaged over seeds and tasks. The proposed methods improve performance individually, suggesting that they all contribute to the final performance. Fp16 crashes and yields no rewards.}
\vspace{-1em}
\label{fig:ablation}
\end{figure}

\subsection{Ablation Experiments}

We additionally perform an ablation experiment to ensure that all our proposed methods improve performance individually. The experiment is structured as follows: we start with the original SAC agent trained in fp16 and then cumulatively add our proposed methods one-by-one. The performance, averaged over environments and seeds, is shown in \Cref{fig:ablation}. As we can see, all proposed techniques improve the performance individually, suggesting that they all are needed to achieve the final result. In \cref{sec:app_more_experiments} we further experiment with removing individual components from the final agent which uses all other modifications and again see that our methods contribute individually to the final performance.

\subsection{Other Numerical Formats}

\label{sec:qtorch}

\begin{figure}
\centering
\includegraphics[width=0.45\textwidth]{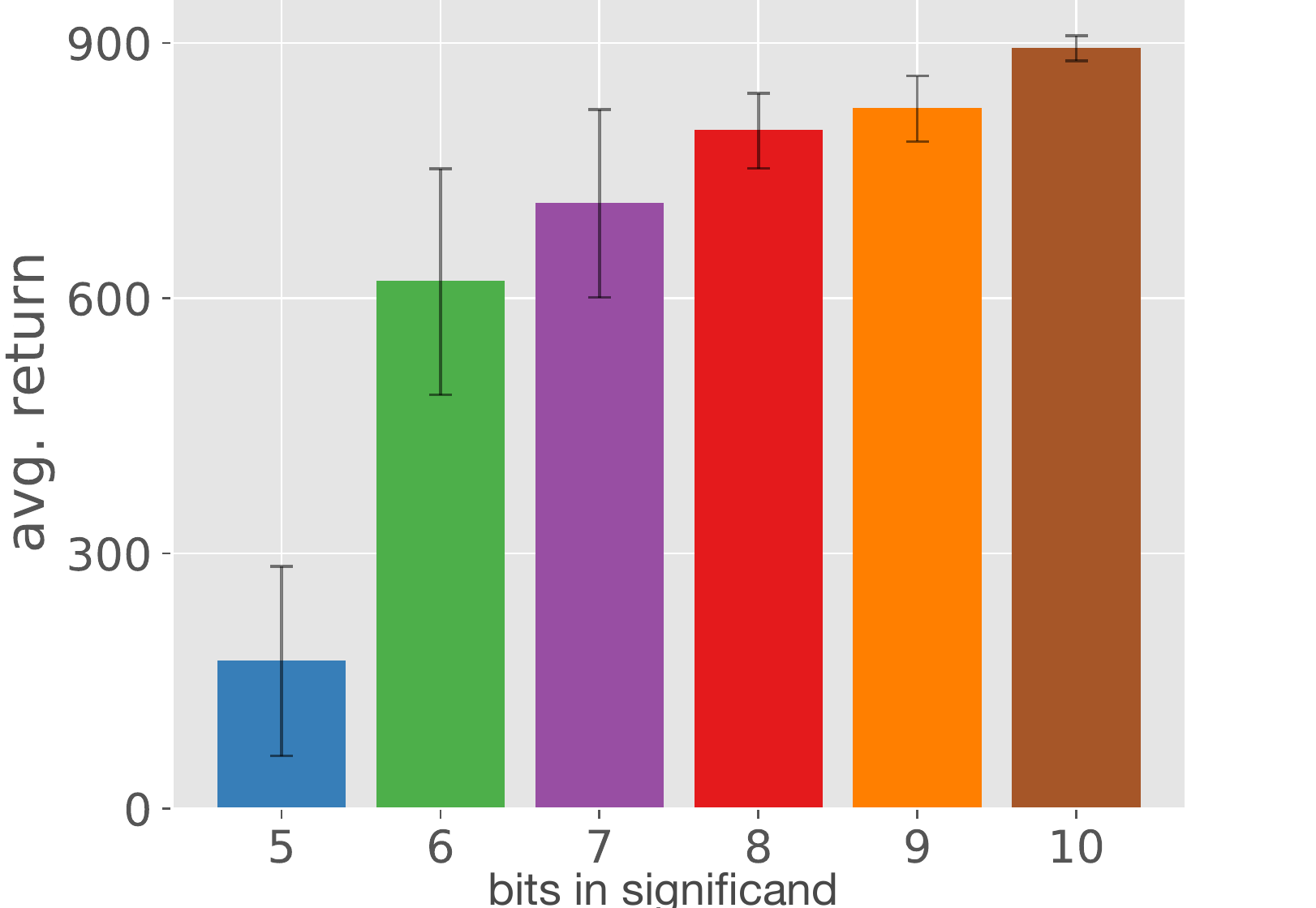}
\caption{Simulation of training in other number formats. The x-axis denotes the number of bits in the significand. Low-precision simulation is performed with qtorch \cite{zhang2019qpytorch}. Results are averaged across tasks. Performance decreases monotonically with fewer bits, first gracefully but then dramatically.}
\vspace{-1.5em}
\label{fig:qtorch}
\end{figure}

In addition to performing experiments in fp16, we also consider training in other numerical formats via the qtorch simulator~\cite{zhang2019qpytorch}. Qtorch allows one to simulate low-precision training by quantizing tensors to some predefined format between calls to the PyTorch backend. Specifically, we start with the fp16 representation and vary the number of bits in the significand while retaining 5 bits for the exponent. For each resulting numerical format, we train SAC with our modifications. Results are shown in~\Cref{fig:qtorch}. We see that the performance degrades with decreased numerical precision, first gracefully but then dramatically when using 5 bits in the significand.

\begin{figure*}
\centering
\includegraphics[width=0.9\textwidth]{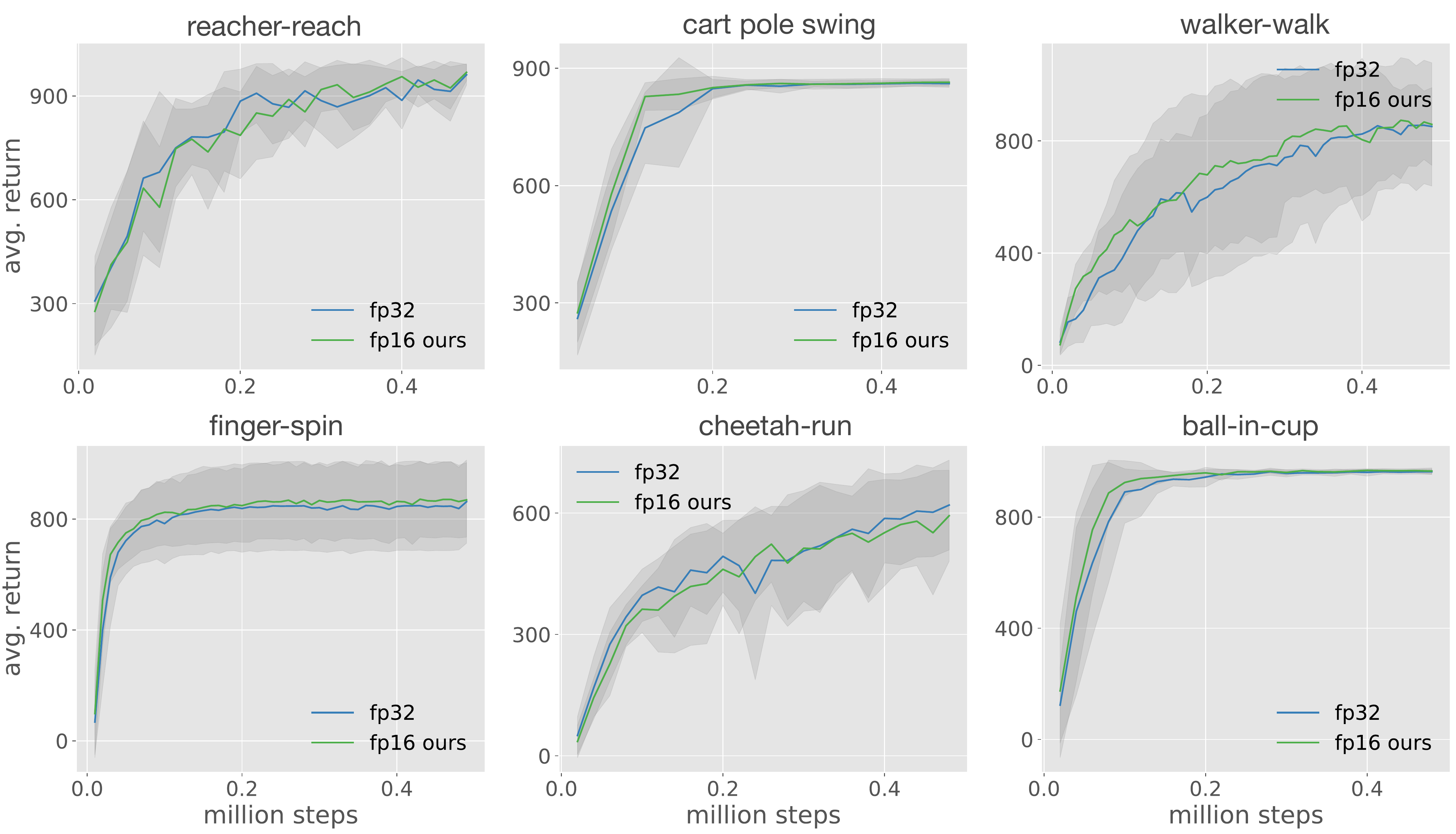}
\vspace{-0.5\baselineskip}
\caption{Learning curves for SAC trained from pixels using standard fp32 training and fp16 training with our modifications. Average performance is close, demonstrating that low-precision RL is feasible from raw images too.}
\vspace{-1em}
\label{fig:from_pixels}
\end{figure*}

\subsection{RL from Pixels}

Until now we have focused on the traditional RL setting of learning from the states of the tasks (i.e. proprioceptive information), which is the original setup for SAC \cite{haarnoja2018soft}. It is also possible to perform RL from raw pixel observations \cite{laskin2020curl, kostrikov2020image}. This setup is more compute-intensive, making low-precision methods very attractive. To this end, we replace the states of our original setup with raw pixel observations processed by a small convolutional network and perform end-to-end RL from pixels. For the convolutional encoder, we follow the general network design of \cite{kostrikov2020image, yarats2019improving, laskin_lee2020rad} -- a few convolutional layers (we use four as \cite{kostrikov2020image}) followed by a linear layer which feeds into a 50-dimensional layer-normalization module whose output is used instead of the environment states. Occasionally we have found the internal variance calculations of layer-normalization to overflow in fp16, to remedy this, we add a weight standardization \cite{qiao2019weight} for the linear layer and downscale its output. This does not change the semantics as layer-normalization is invariant under rescaling. The issue could possibly be addressed at the GPU kernel level, but since the cudnn kernels are not open source we defer such studies to future work. Images are resized to 84-by-84 tensors and we follow \cite{kostrikov2020image} regarding frame-stacking, image augmentation and hyperparameters (importantly learning rate is increased to $1\mathrm{e}{-3}$), see \Cref{sec:appendix_conv} for details.

In \Cref{fig:from_pixels}, we compare the performance between training from pixels in fp32 and fp16, the latter using our modifications from \cref{sec:proposed_method}. Again, we see that the performance is very close, despite one network using half-precision numbers. This demonstrates the feasibility of RL from pixels in low precision too, where high-dimensional image input can result in compute-intensive workloads.

\subsection{Performance}

We now measure the performance improvements obtained from low-precision training, varying the batch size and the width (i.e. number of filters) of the convolutional layers to simulate different computational demands. We use all our proposed methods for low-precision training, larger improvements can likely be obtained for environments that work well without all methods. We here focus on RL from pixels, \Cref{sec:appendix_perf} contains the measurements for learning from states. When training the agent, there are three main components of the compute time: 1) simulating the environment, 2) sampling batches and 3) GPU computation. To achieve optimal performance, these operations should be asynchronous; however, the codebase of \cite{pytorch_sac} is synchronous. As asynchronous learning is well-understood and not the focus of our work, we only measure the compute time of the third component, using one fixed batch and performing multiple training steps. The time is averaged over 500 gradient updates (and occasional momentum updates for the target network) with a warm started Tesla V100 GPU. We perform measurements with PyTorch tools, see \Cref{sec:appendix_perf} for tooling details.

\begin{table}[]
\centering
\caption{Time (milliseconds) for processing one minibatch as a function of network width and batch size (bsize). As computational demands increase the improvement from fp16 training exceeds a factor of two. Measurements are performed on the Cheetah task.}
\label{tab:compute}
\begin{tabular}{lllll}
\toprule
width / bsize &  32/512 & 32/1024 & 64/512 & 64/1024 \\ \midrule
fp32 & 92.98 & 181.53 & 188.96 & 373.43 \\
fp16 (ours) & 76.46 & 127.10 & 93.76 & 171.12 \\ \midrule
improvement & 1.22 & 1.43 & 2.02 & 2.18 \\ \bottomrule
\end{tabular}
\vspace{-1.2em}
\end{table}

\begin{table}[]
\centering
\caption{Memory (GB) consumed as a function of network width and batch size (bsize). Across all batch sizes and network widths savings are close to a factor of two. Measurements are performed on the Cheetah task.}
\label{tab:memory}
\begin{tabular}{lllll}
\toprule
bsize/ \# filters & 512/32 & 1024/32 & 512/64 & 1024/64 \\ \midrule
fp32 & 2.55 & 4.94 & 4.23 & 8.21 \\
fp16 (ours) & 1.36 & 2.61 & 2.28 & 4.37 \\ \midrule
improvement & 1.87  & 1.89 & 1.86 & 1.88 \\ \bottomrule
\end{tabular}
\vspace{-1.2 em}
\end{table}

In \Cref{tab:compute}, we compare the time per update for different network widths and batch size. We see that the speedup increases to more than 2x as the computational demands increase. The reason that performance changes is likely that the smallest configuration does not saturate available compute resources of the Tesla V100, however factors such as memory hierarchy and pipeline balance might also play a role. In \Cref{tab:memory}, we show the memory improvements of training in fp32 vs. fp16. The memory improvements stay around 1.87x for all network widths, demonstrating dramatic improvements that can enable larger models on edge devices and the small memory overhead of our methods.

\subsection{Investigating Gradient Scales}

\begin{figure}
\centering
\includegraphics[width=0.5\textwidth]{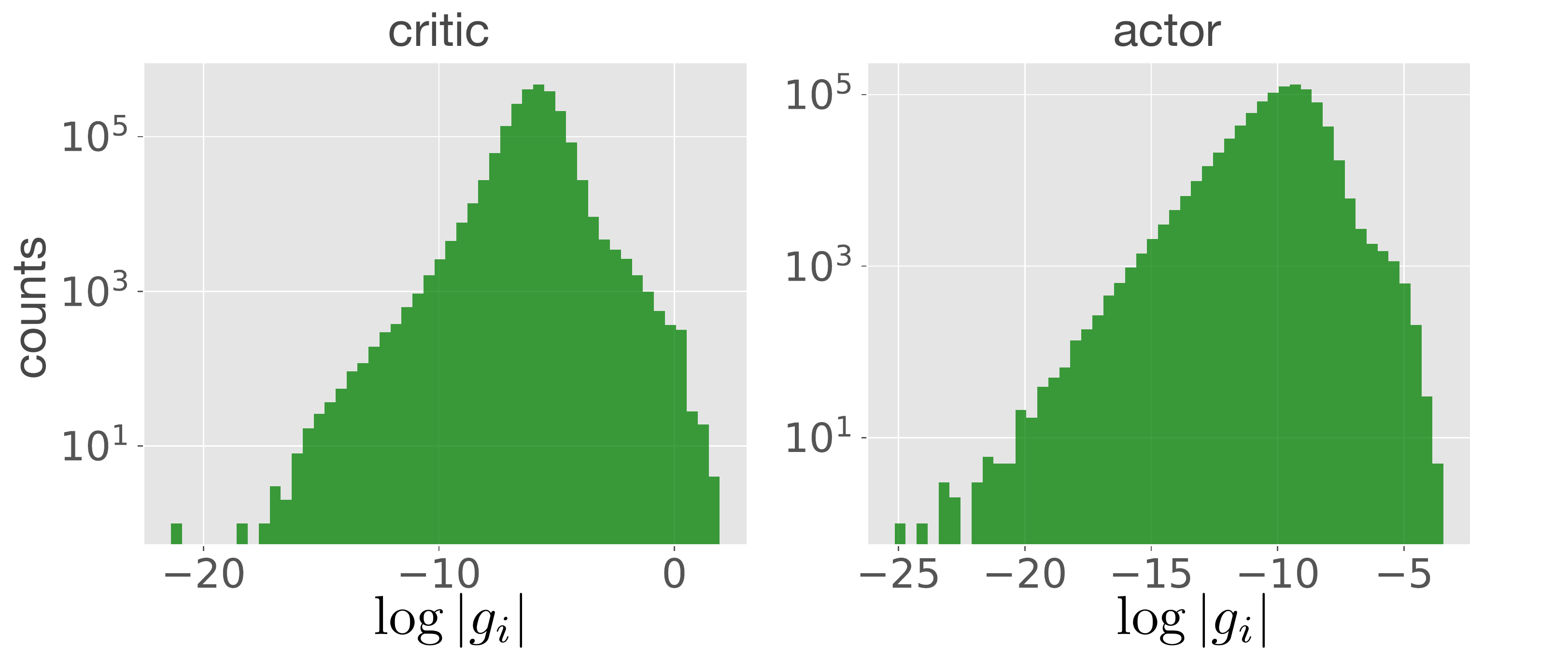}
\vspace{-1.5\baselineskip}
\caption{Histogram over the gradients for the critic and actor networks. Note that both the x-axis and the y-axis are logarithmic. The gradients have a dynamic range spanning orders of magnitude, which can be problematic to represent in low precision.}
\vspace{-1.5em}
\label{fig:grad_scale}
\end{figure}

The good performance of our proposed methods indicates that the numerical stability of policy parametrization, precision in gradient accumulation, and the dynamic range of the gradients are important issues in low-precision RL. We here seek to illustrate the wide gradient distribution directly. In \Cref{fig:grad_scale}, we show the distribution of gradients for the cheetah task (trained in fp32) after 250,000 environment steps. Note that both axes use a logarithmic scale. The gradients have many orders of magnitude of dynamic scale, and when Adam computes the square of these even more dynamic range is needed. This suggests that gradient scale is an important issue in low precision RL.

\subsection{Further Experiments}

We have demonstrated that our methods can match fp32 performance for the tuned default SAC parameters of \citet{pytorch_sac}. As we did not tune these parameters ourselves, controlling for effects of hyperparameter tuning is not straightforward \cite{jordan2020evaluating}. To demonstrate that our methods are stable for different hyperparameters, we consider experiments with random hyperparameters in \Cref{sec:app_rnd_params}. These show that our method still matches fp32 rewards for such random parameters, thus demonstrating the parameter stability of our methods.

%% file: text/related_concl.tex
\section{Related Work}

\textbf{Accelerating RL.} There are many proposed approaches for accelerating RL. \cite{mnih2016asynchronous, babaeizadeh2016reinforcement} demonstrate the efficacy of asynchronous training, whereas distributed training has received more attention lately~\cite{horgan2018distributed, espeholt2018impala, wijmans2019dd, espeholt2019seed, kapturowski2018recurrent}. Distributed RL training has received support from the Rllib project~\cite{moritz2018ray, liang2017rllib}, other notable codebases are Rlpyt \cite{stooke2019rlpyt} and torchbeast \cite{kuttler2019torchbeast}. Additionally, \cite{petrenko2020sample} and \cite{liu2020high} have recently demonstrated algorithms for high throughput RL on a single compute node. \cite{dalton2019accelerating} has investigated simulating the environment on the GPU for acceleration. \cite{lau2002adaptive} proposed vector quantization for traditional tabular reinforcement learning -- i.e., with no deep learning component. The only work on accelerating deep RL with low precision known to us is the pre-print of \cite{krishnan2019quantized}. It differs from and/or complements our work by 1) studying post-training quantization, which is not the focus of our work, 2) using mixed precision, which we experimentally demonstrate does not work in general .

\textbf{Low-Precision Deep Learning.} Low-precision deep learning goes back at least to \cite{gupta2015deep}, which has inspired large amounts of subsequent work. Among the most noteworthy approaches is mixed precision \cite{micikevicius2017mixed}, where full and half precision are used together. Such settings often necessitate loss scaling to avoid the gradients underflowing in the backward pass. Furthermore, it requires a larger set of instructions at the hardware level and also moving parameters between e.g. fp32 and fp16 representations. Other relevant approaches include binary weights~\cite{courbariaux2015binaryconnect}, fixed point~\cite{lin2016fixed}, B-float \cite{kalamkar2019study}, block floating point~\cite{song2017computation}. 8-bit~\cite{wang2018training}, low-precision accumulation \cite{ni2020wrapnet} or integer arithmetic~\cite{jacob2018quantization, das2018mixed}. Another strategy is to average the low-precision weights into full precision \cite{yang2019swalp} or to perform stochastic rounding \cite{hopkins2020stochastic}. For recent theoretical work on quantization, see \cite{chen2020statistical}. Kahan summation has recently been proposed in deep learning for gradient accumulation \cite{zamirai2020revisiting, han2019auto}, but not for momentum updates as we do. Post-training quantization is also a popular strategy \cite{fang2020post}, but not the focus here.

\section{Conclusions}

In this paper, we demonstrate the feasibility and potential performance gains of low-precision RL. Whereas na\"ive strategies from supervised learning fail we propose six modifications to the SAC agent \cite{haarnoja2018soft} which improve numerical stability. With our modifications, fp16 training matches full precision rewards while improving memory footprint and compute time. Future directions include mixed precision and extending our methods to more agents. We hope that our methods and code provide a starting point for researchers interested in low-precision RL.

\section*{Acknowledgements}

This research is supported in part by the grants from the National Science Foundation (III-1618134, III-1526012, IIS1149882, IIS-1724282, and TRIPODS- 1740822), the Office of Naval Research DOD (N00014- 17-1-2175), Bill and Melinda Gates Foundation. We are thankful for generous support by Zillow and SAP America Inc. This material is based upon work supported by the National Science Foundation under Grant Number CCF-1522054. We are also grateful for support through grant AFOSR-MURI (FA9550-18-1-0136). This work was supported by a gift from SambaNova Systems, Inc. Any opinions, findings, conclusions, or recommendations expressed here are those of the authors and do not necessarily reflect the views of the sponsors. We thank Rich Bernstein, Yiwei Bai, Wenting Zhao and Ziwei Liu for help with the manuscript. We also thank David Eriksson for pointers regarding numerics.

%% file: text/appendix.tex
\section{Ethics Statement}

Deep learning models have grown extremely large over the last few years, and their computational demands keep on increasing. This translates into substantial energy consumption which can have environmental consequences. By performing model training and inference in low precision, it becomes possible to reduce energy expenditure. Deep reinforcement learning methods have many practical applications and can quickly transform into products such as domestic robots and self-driving cars. Upon commercialization, low-precision RL could enable substantial energy savings. RL has the potential for military and malicious applications, but also positive societal outcomes such as scientific discovery and autonomous vehicles. We do not perceive that our work differs significantly from other RL work in this regard.

\section{Experimental details}
\label{sec:hyperparams}

The hyper-parameters for SAC, following \cite{pytorch_sac}, are given in \Cref{tab:sac_hp}. We also use the network architecture of \cite{pytorch_sac}: both the actor and critic networks have hidden depth 2 and hidden dimensions 1024. The actor outputs $\log \sigma$ for the actions, which is coerced to lie in $[-5, 2]$ via a $\tanh$ non-linearity. The hyperparameters of our methods are listed in \cref{tab:our_hp}. For dynamic loss scaling, we follow the strategy of PyTorch amp~\cite{amp}. The scale is initialized at some large value $init\_grad\_scale$. After each backward pass, we inspect the gradients. If there are non-finite values, we decrease the loss scale by a factor of 2. If we observe no such issues for $inc\_grad\_scale\_freq$ consecutive epochs, we increase the scale by a factor of 2 and reset this counter. To choose a proper $K$ in \cref{eq:softplus_fixed}, one needs to consider $M_{\max}$, the largest number that can be represented in the numerical format. To avoid overflow of $\log(1 + \exp(x))$ in the backwards pass in a naive implementation, one should exchange it for a linear function for $x \approx \log M_{\max}$. Larger limits are possible depending on how the soft-plus function is implemented, we take $10$ as it is a round number and works well in practice. The Kahan-momentum updates are made to a scaled buffer to avoid underflow as $\tau$ can be small, we use a scale of $1e4$.

\begin{table}[h]
\caption{Hyper-parameters used for SAC, following \cite{pytorch_sac}.}
\label{tab:sac_hp}
\begin{center}
    \begin{tabular}{ l | l }
    \hline
    Parameter & Value \\ \hline
    $\gamma$ & 0.99 \\ 
    $T_0$ & 0.1 \\ 
    $\tau$ & 0.005 \\
    $\alpha_{\text{adam}}$ & 1e-4 \\
    $\epsilon_{\text{adam}}$ & 1e-8 \\
    $\beta_1{\text{adam}}$ & 0.9 \\
    $\beta_2{\text{adam}}$ & 0.999 \\
    batch size & 1024 \\
    target update freq & 2 \\
    seed steps & 5000 \\
    log $\sigma$ bounds & [-5, 2] \\
    actor update frequency & 1 \\
    \hline
    \end{tabular}
\end{center}
\end{table}

\begin{table}[h]
\caption{Hyper-parameters for our methods.}
\label{tab:our_hp}
\begin{center}
    \begin{tabular}{ l | l }
    \hline
    Parameter & Value \\ \hline
    $init\_grad\_scale$ & 1e4 \\
    $inc\_grad\_scale\_freq$ & 1e4 \\
    $K$ & 1e1 \\
    Kahan-momentum scale  & 1e4 \\
    \hline
    \end{tabular}
\end{center}
\end{table}

\section{Proof of \Cref{th:main}}
\label{sec:appn_proof}

\textbf{hAdam.} We prove this by induction on the statement $w_t = \sqrt{v_t}$. It holds at $t=0$ as both buffers are initialized to zero, thus we only need to prove the induction step. Assuming $w_t=\sqrt{v_t}$ we have

$$
w_{t+1} =  \text{hypot}( \sqrt{\beta} w_t, \sqrt{(1-\beta)} g_{t+1})
$$

$$
= \text{hypot}( \sqrt{\beta} \sqrt{v_t}, \sqrt{(1-\beta)} g_{t+1})
$$

$$
 = \sqrt{\beta v_t + (1-\beta) g_{t+1}^2} = \sqrt{v_{t+1}}
$$

For the Adam/hAdam updates we then have $\frac{m}{\sqrt{v} + \epsilon} = \frac{m}{w + \epsilon}$, and thus the updates are identical.

\textbf{Compound loss-scaling.} For any $\gamma > 0$ we have $ \frac{\gamma m}{\gamma (w + \epsilon)} = \frac{ m}{ w + \epsilon}$. Thus, as long as $\epsilon$ is scaled by $\gamma$, this modification changes nothing in infinite precision. 

\textbf{Normal-fix.} In infinite precision we have $\frac{(x-\mu)^2}{\sigma^2} = \big( \frac{x-\mu}{\sigma} \big)^2 $. Thus, the normal fix will not change anything.

\textbf{Kahan-momentum.} In infinite precision where arithmetic is commutative and associative, Kahan summation reduces to normal summation.

\textbf{Kahan-gradients.} In infinite precision where arithmetic is commutative and associative, Kahan summation reduces to normal summation. $\blacksquare$

\section{Infrastructure}
\label{sec:infra}

Experiments were conducted with PyTorch 1.7.0 on Nvidia Tesla V100 GPUs using CUDA 10.2 and CUDNN 7.6.0.5, except for the performance measurements where we also consider CUDA 11.0 and CUDNN 8.0.0.5. See \Cref{sec:appendix_perf} for details on the performance measurements.

\section{Additional Experiments}

\label{sec:app_more_experiments}

\begin{figure}[h]
\centering
\includegraphics[width=0.5\textwidth]{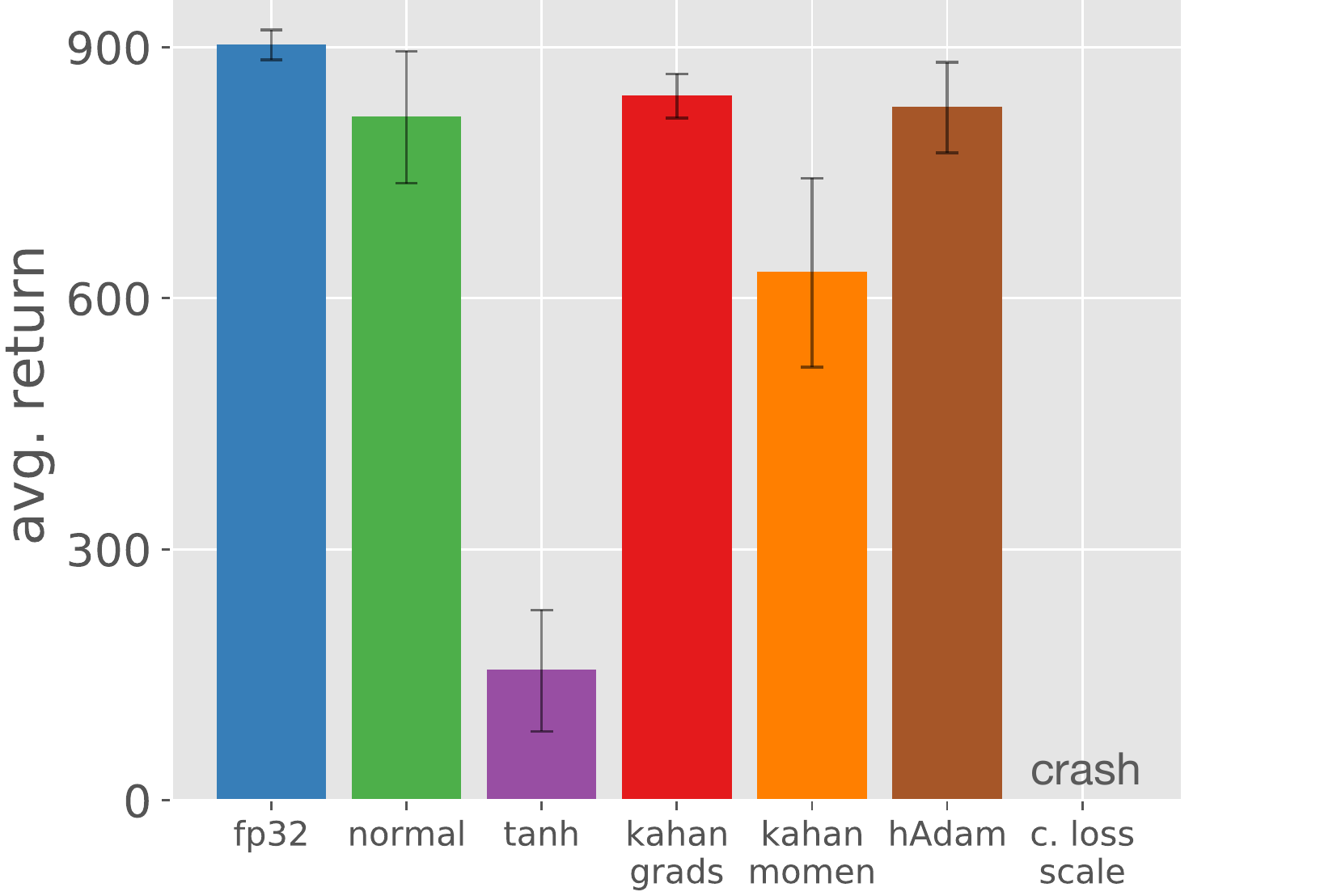}
\caption{We remove one of our proposed method from the complete agent (which uses the other 5 methods) and then train the agent in fp16. Results are averaged over seeds and environments. Removing any single method decreases the performance, suggesting that all proposed methods are needed to reach the final performance.}
\label{fig:loo_ablation}
\end{figure}

\begin{figure*}[h]
\centering
\includegraphics[width=\textwidth]{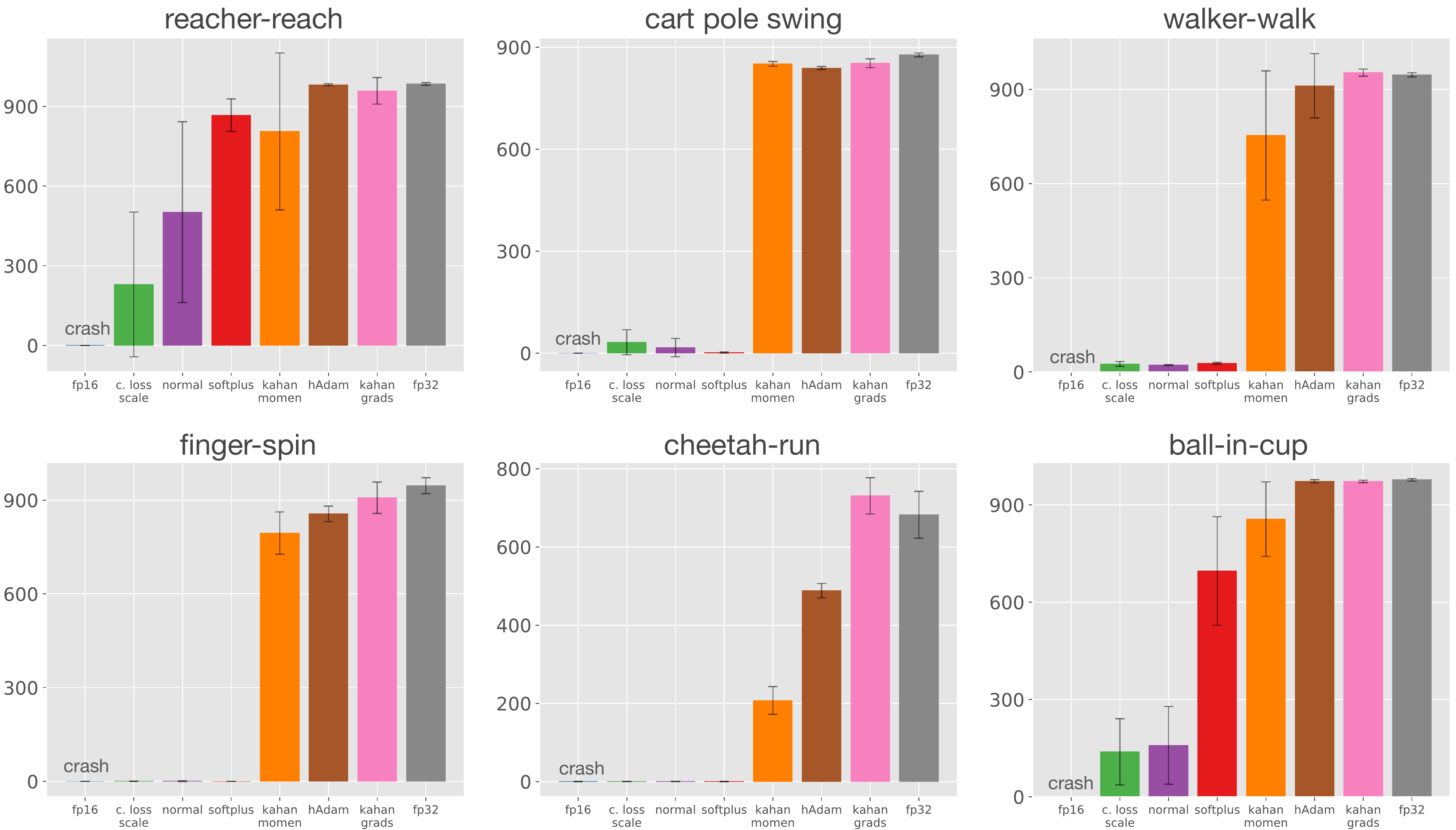}
\caption{Performance of fp16 training as we add our proposed methods one by one. Scores are broken down by individual tasks. For all tasks, many proposed methods are needed to reach satisfactory performances. However, the games differ in how many are needed, suggesting that some tasks might be more numerically robust. This figure follows \Cref{fig:ablation}, but for individual tasks.}
\label{fig:ablation_per_game}
\end{figure*}

We here present some additional supporting experiments. In \Cref{fig:ablation_per_game} we show the original ablation experiments with results broken down by task. We see that all tasks require several of the proposed methods to work well, but also variation between tasks. Specifically, it seems like some tasks require fewer methods to work well -- suggesting that these are more robust to numerical issues. To further verify that our proposed methods contribute individually to the performance, we perform an ablation experiment where we remove one component from the final agent which uses all other techniques. The results, averaged over seeds and environments, are shown in \Cref{fig:loo_ablation}. We see that all proposed methods are needed to reach satisfactory performance across all games.

\begin{figure}[h]
\centering
\includegraphics[width=0.5\textwidth]{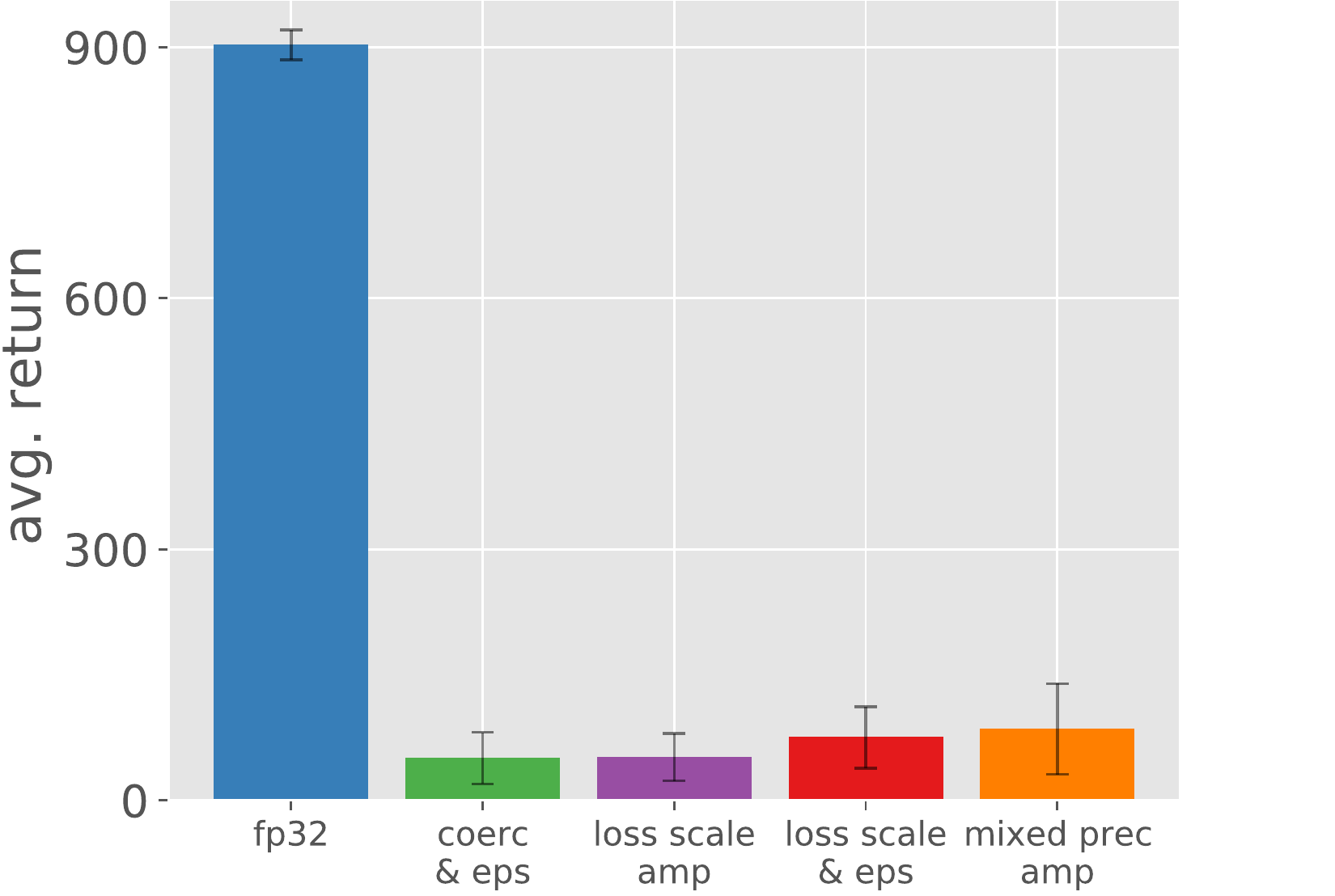}
\caption{The performance of our methods compared to a few additional baselines. We specifically consider 1) standard loss scaling, but using the amp default settings for the dynamic loss scaling, referred to as amp; and 2) increasing the value of $\epsilon$ in Adam by a factor of 10 to increase the numerical stability, referred to as eps. None of these methods work well.}
\label{fig:additional baseline}
\end{figure}

We also compare our proposed method against the baselines from the main text with some hyperparameter modifications. Specifically, we consider 
1) using the default amp settings for the loss scaler schedule -- an initial scale of $2^{16}$ and a growth interval of $2000$. We refer to this modification as amp in the figures. We also consider 
2) to increase $\epsilon$ in Adam by a factor of 10 to stabilize training. We refer to this modification as eps in the figures. Results are shown in \Cref{fig:additional baseline}. None of these methods improve the training substantially.

\begin{figure*}
\centering
\includegraphics[width=0.9\textwidth]{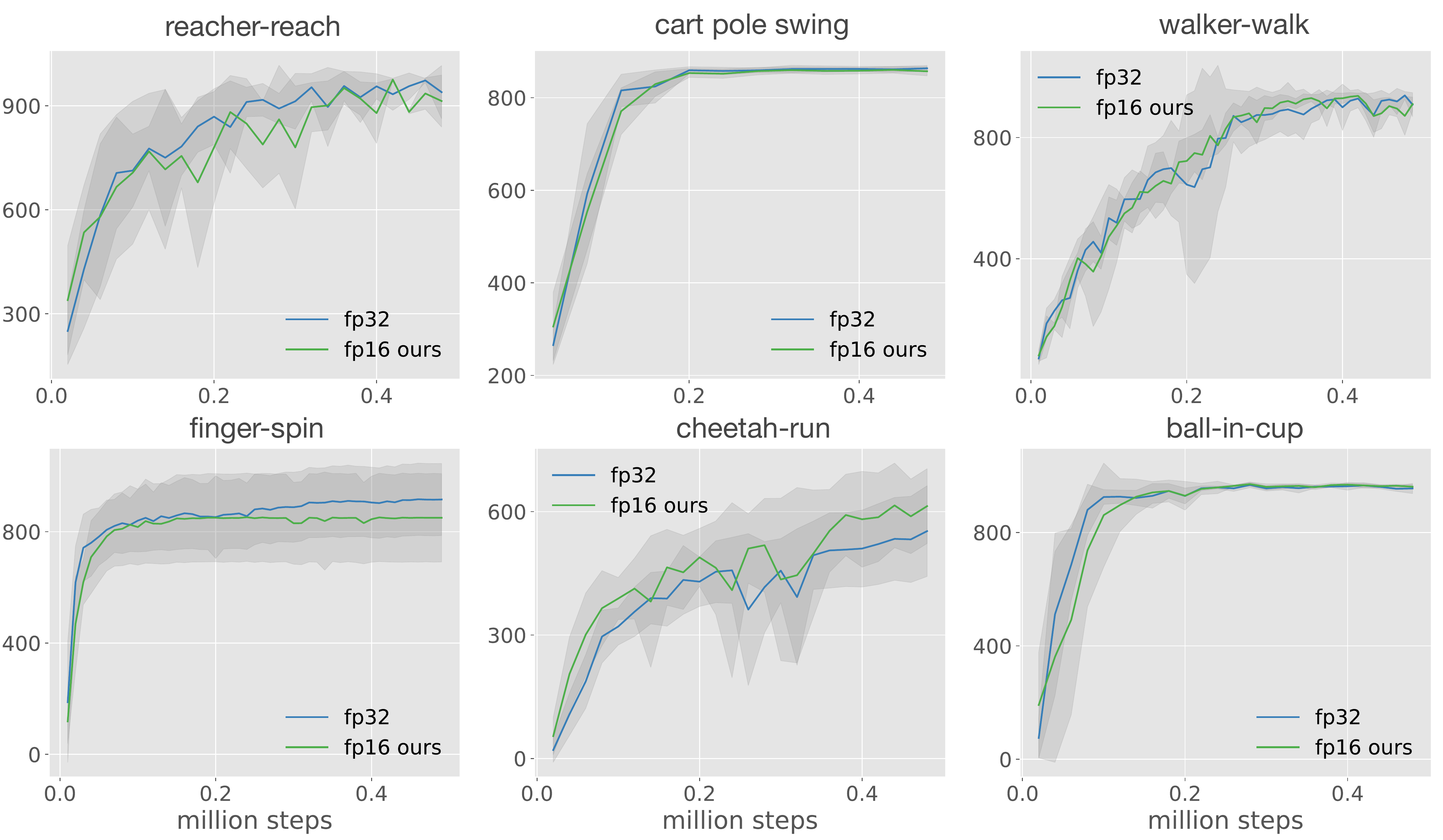}
\vspace{-0.5\baselineskip}
\caption{Learning curves for SAC trained from pixels comparing standard fp32 training and training in fp16 with our modifications. The fp32 baseline here does not use weight standardization. Average performance is again close.}
\label{fig:planet_cnn_nows}
\vspace{-1em}
\end{figure*}

In the main text, we compare against training RL from pixels with weight standardization applied to the linear layer. In \Cref{fig:planet_cnn_nows} we compare our method (which uses weight standardization) against an fp32 baseline which does not. Again, the results are close.

\section{Random Parameters}

\label{sec:app_rnd_params}

Most RL algorithms requires hyperparameter tuning to successfully adapt to new environments. In the main text, we simply use the default hyperparameters from \cite{pytorch_sac}. To demonstrate the parameter stability of our method, we now consider random hyperparameters. We generate five sets of random hyperparameters, randomizing learning rate $\alpha$, discount $\gamma$, initial temperature $T_0$, batch size, critic update speed $\tau$, action bound as per \Cref{tab:rnd_params_dis}, keeping other parameters from \cite{pytorch_sac} (see \Cref{tab:sac_hp}). The average scores are given in \Cref{tab:rnd_params}, and our method performs close to fp32 across the different hyperparameter groups, thus showing that our method is stable across hyper parameters.

\newcommand{\expnumber}[2]{{#1}\mathrm{e}{#2}}

\begin{table*}[]
\begin{center}
\caption{Random hyperparameters obtained. Learning rate is obtained from a log uniform distribution over $[\expnumber{1}{-5}, \expnumber{1}{-3}]$, $\min \log \sigma$ over a uniform distrubution over $[-7, -3]$, $\tau$ over a uniform distribution over $[0.0025, \expnumber{1}{-2}]$, $T_0$ over a log uniform distribution over $[\expnumber{1}{-2}, \expnumber{1}{-1}]$ and batch size from the discrete distribution $\{512, 2024, 2048\}$.}
\label{tab:rnd_params_dis}
\begin{tabular}{lllllll}
\toprule
 & $\gamma$ & learning rate & $\min \log \sigma$ & $\tau$ & $T_0$ & batch size \\
 \midrule
params 1 & 0.921 & 0.000669 & -5.37 & 0.00331 & 0.0797 & 512 \\
params 2 & 0.983 & 0.0000751 & -5.64 & 0.00289 & 0.0137 & 2048 \\
params 3 & 0.979 & 0.0000751 & -6.99 & 0.00577 & 0.0122 & 2048 \\
params 4 & 0.982 & 0.00097 & -4.07 & 0.00327 & 0.0111 & 2048 \\
params 5 & 0.948 & 0.0000263 & -6.92 & 0.00933 & 0.14 & 1024 \\
\bottomrule
\end{tabular}
\end{center}
\end{table*}

\begin{table*}[h]
\caption{Average scores for random hyperparameters. Scores are averaged across 3 seeds and 6 environments. Parameters are generated per \Cref{tab:rnd_params_dis}. The scores are similar for fp32 and our fp16 agent, demonstrating that our methods can match fp32 results robustly across parameters.}
\label{tab:rnd_params}
\begin{center}
    \begin{tabular}{ l l l l l l }
    \toprule
    avg. reward & params 1 & params 2 & params 3  & params 4  & params 5  \\ 
    \midrule
    fp32 & 767 $\pm$ 11 & 877 $\pm$ 14  & 872 $\pm$ 12   & 887 $\pm$ 60   & 732 $\pm$ 50   \\
    fp16 (ours) & 778 $\pm$ 27  & 869 $\pm$ 33  & 862 $\pm$ 29   & 880 $\pm$ 54   & 709 $\pm$ 36  \\
    \bottomrule
    \end{tabular}
\end{center}
\end{table*}

\section{Details on RL from Pixels}
\label{sec:appendix_conv}

The hyper-parameters we use for SAC from pixels follow \cite{kostrikov2020image}. They are largely the same as for SAC from states, the differences are listed in \Cref{tab:sac_cnn_hp}. We use 100 as the scale of the Kahan momentum and add a $\epsilon$ ($1\mathrm{e}{-4}$) to the outputted $\sigma$ of the actor. This prevents underflow which now can happen as \cite{kostrikov2020image} uses a larger range for the stds. Otherwise, we use hyper-parameters from \Cref{tab:our_hp} for our methods. Following \cite{hafner2019learning, kostrikov2020image} we use action repeat for the tasks as per \Cref{tab:repeat}. Images are resized to 84-by-84 RGB images, we then use frame stacking of three to get input shapes of (9, 84, 84). We follow \cite{kostrikov2020image} and apply image augmentations to the input. Specifically, we use random cropping with padding by 4 and input both an augmented batch of size 512 and an original batch of 512 to the critic network, resulting in a total batch size of 1024. The actor and $\alpha$ uses a batch size of 512 for their updates.

For the CNN encoder, we use four convolutional layers with ReLu non-linearity between them. All convolutional layers have spatial extent 3-by-3, the first layer uses a stride of 2, the others use stride 1. By default, all convolutional layers have 32 filters. The feature map from the convolutions is fed into a linear layer which outputs a 50-dimensional vector which is fed into a layer-normalization layer \cite{ba2016layer}. We occasionally observed the internal variance calculations of the layer normalization overflowing. To remedy this, we apply weight standardization \cite{qiao2019weight} to the linear layer and further down-scale output larger than 10 to 10, this avoids overflow in the layer-norm calculations. Since layer-norm is invariant under rescaling the input and adding a constant term, these modifications will not change layer-norm in infinite precision. This strategy could likely be implemented in the layer-norm CUDA kernel, but since it is not open-source we defer such investigations to future work. Please consult our code for the PyTorch implementation which is based upon the codebase of \cite{kostrikov2020image}.

\begin{table}[h]
\centering
\begin{tabular}{l|c}
\toprule
task & action repeat \\
\midrule
Cartpole Swingup &  $8$ \\
Reacher Easy &  $4$ \\
Cheetah Run & $4$ \\
Finger Spin & $2$ \\
Ball In Cup Catch & $4$ \\
Walker Walk & $2$ \\
\bottomrule
\end{tabular}
\caption{The action repeat hyper-parameter for each task, values come from \cite{hafner2019learning}.}
\label{tab:repeat} 
\end{table}

\begin{table}[h]
\caption{Hyper-parameters used for SAC from pixels, following \cite{kostrikov2020image}. We only list hyperparameters that differ from those given in \cref{tab:sac_hp}.}
\label{tab:sac_cnn_hp}
\begin{center}
    \begin{tabular}{ l | l }
    \hline
    Parameter & Value \\ \hline
    $\tau$ & 0.01 \\ 
    $\alpha_{\text{adam}}$ & 1e-3 \\
    seed steps & 1000 \\
    actor update frequency & 2 \\
    log $\sigma$ bounds & [-10, 2] \\
    \hline
    \end{tabular}
\end{center}
\end{table}

\section{Details on Performance Measurements}
\label{sec:appendix_perf}

Memory consumption and throughput are measured and averaged over 500 iterations, with 500 iterations as a warm start. We include momentum updates for the target network in these measurements. Time is measured with CUDA events, which are provided natively in PyTorch. Memory is simply measured as the maximum CUDA memory allocated during training, which again is provided natively in PyTorch. We observed non-deterministic CUDNN sometimes settling for sub-optimal GPU kernels. To provide the memory and compute measurements closest to optimal performance, we report the best numbers obtained when setting CUDNN to deterministic or non-deterministic and when using 1) CUDA 11.0 and CUDNN 8.0.0.5 or 2) CUDA 10.2 and CUDNN 7.6.0.5.

In \Cref{tab:states_compute} we show the performance of SAC from states and observe that the gains become very large as the computational demands grow. Performance differences for the largest models might be related to cache issues rather than the absolute improvement in speed for individual floating-point operations. For the smallest models, where updates take less than 20 milliseconds, the overhead of our methods is larger than the gains from using half-precision numbers. Again, the reason for this is likely that such small workloads do not saturate available parallel resources on the V100, for less performant hardware we would expect a larger difference for small configurations. In \Cref{tab:states_memory} we instead show the memory footprint for SAC from states. The improvement is relatively constant but scales somewhat differently with the batch size of model capacity. This is natural as the memory footprint of Kahan summation scales with model size. We have observed that moving the entire network to half-precision, without any further modifications, can result in memory savings slightly under $50 \%$. Thus, it seems like the CUDA kernels or the tensor caching policy is not identical for fp32 and fp16. If they were, we would likely see slightly larger memory improvements.

\begin{table}[]
\centering
\caption{Time (milliseconds) for processing one minibatch for SAC from states as a function of network width and batch size (bsize), measured for the Cheetah task. For the smallest model, where a update takes less than 20 milliseconds, the overhead of our fp16 method is larger than the gains. However, as the computational demands increase the benefits grow very large.}
\label{tab:states_compute}
\begin{tabular}{lllll}
\toprule
width & 1024 & 1024 & 4096 & 4096 \\ 
bsize & 1024 & 4096 & 1024 & 4096 \\ \midrule
fp32 & 16.63 & 17.94 & 58.22 & 202.38 \\
fp16 (ours) & 17.38 & 16.99 & 20.58 & 45.64 \\ \midrule
improvement & 0.96 & 1.06 & 2.83  & 4.43  \\ \bottomrule
\end{tabular}
\vspace{-1.2em}
\end{table}

\begin{table}[]
\centering
\caption{Memory (MB) consumed for SAC from states as a function of network width and batch size (bsize), measured for the Cheetah task. The memory benefits are relatively constant across computational demands, but scale slightly differently with batch size and model capacity. This is natural as Kahan summation requires storing tensors the size of the model weights.}
\label{tab:states_memory}
\begin{tabular}{lllll}
\toprule
width & 1024 & 1024 & 4096 & 4096 \\ 
bsize & 1024 & 4096 & 1024 & 4096 \\ \midrule
fp32 & 128 & 320 & 1265  & 1973 \\
fp16 (ours) & 77 & 185 & 826 & 1163 \\ \midrule
improvement & 1.67   & 1.73  & 1.53  & 1.7  \\ \bottomrule
\end{tabular}
\vspace{-1.2 em}
\end{table}

\section{Comparison of Learned Models with Different Numerical Precision}
\label{sec:appendix_prec}

\begin{figure}[!htbp]
    \centering
    \subfigure[]{
        \includegraphics[width=0.48\textwidth]{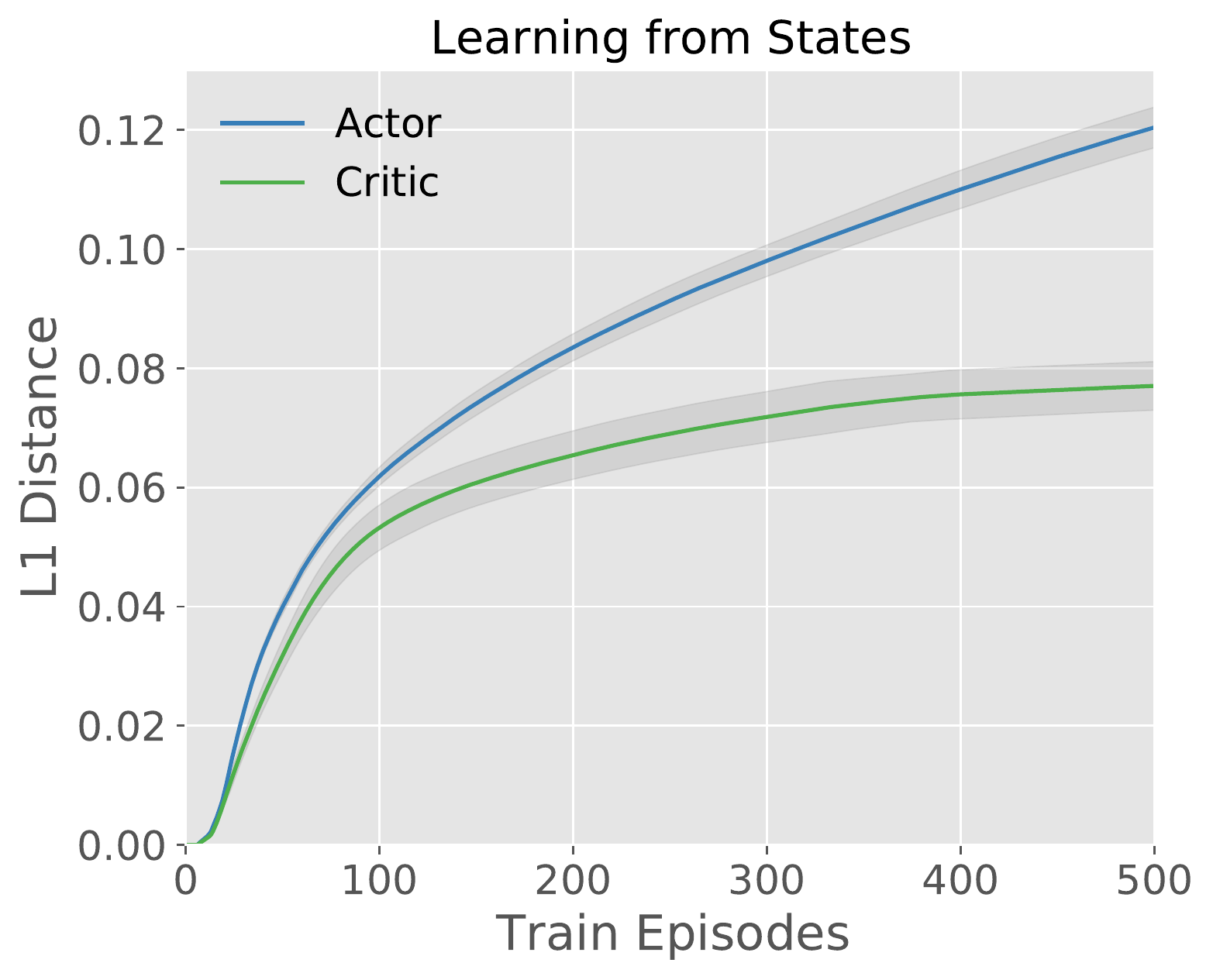}
    }
    \subfigure[]{
        \includegraphics[width=0.48\textwidth]{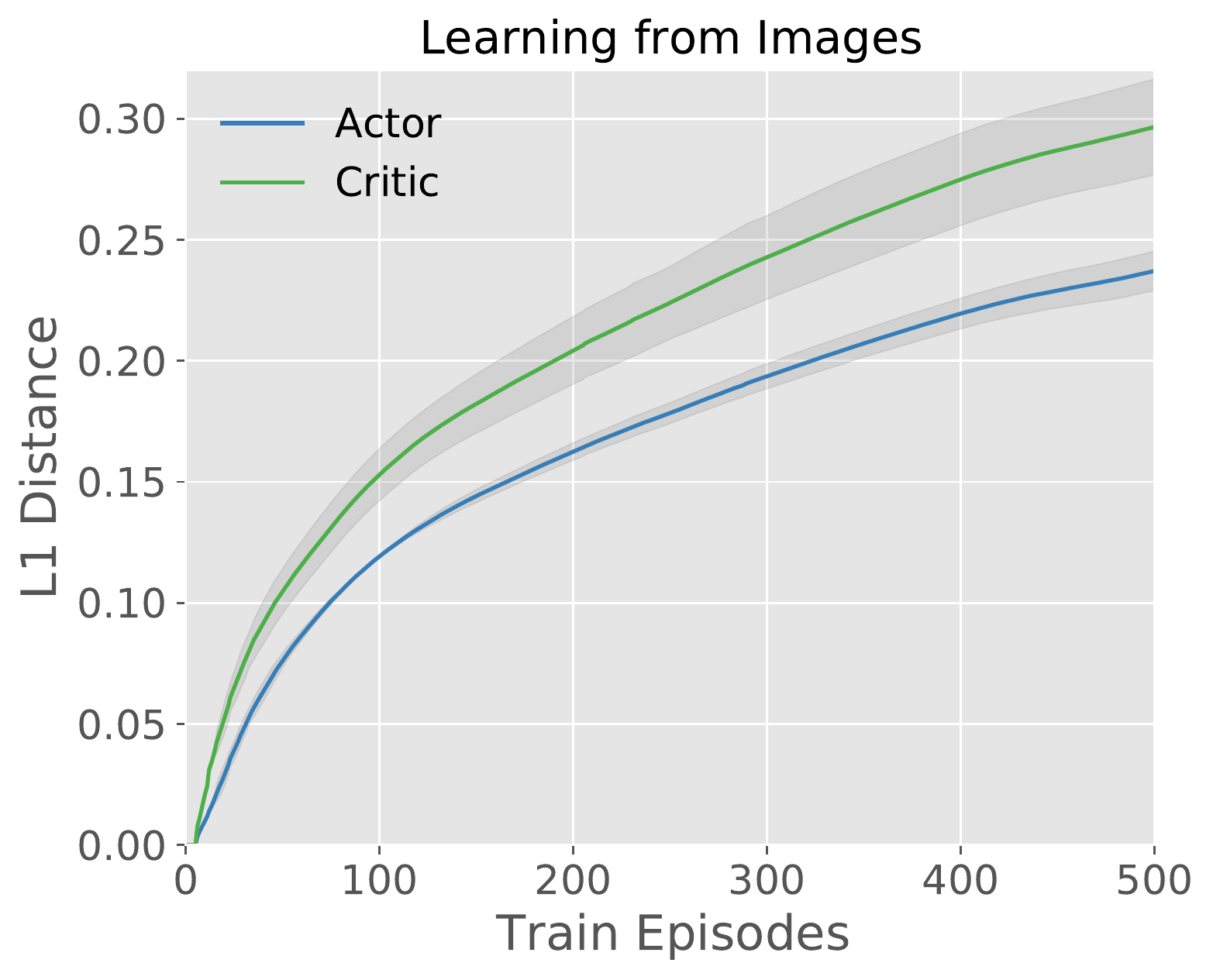}
    }
    \caption{Average L1 distance between the actor/critic weights during training. We compare in 2 settings: (a) \emph{learning from states} and (b) \emph{learning from images}. We trained 3 pairs of SAC agents, each pair with the same random seed, and report the L1 distance between learned weights averaged over 3 pairs.}
    \label{fig:weight_diff}
\end{figure}

\begin{figure}[!htbp]
    \centering
    \includegraphics[width=0.48\textwidth]{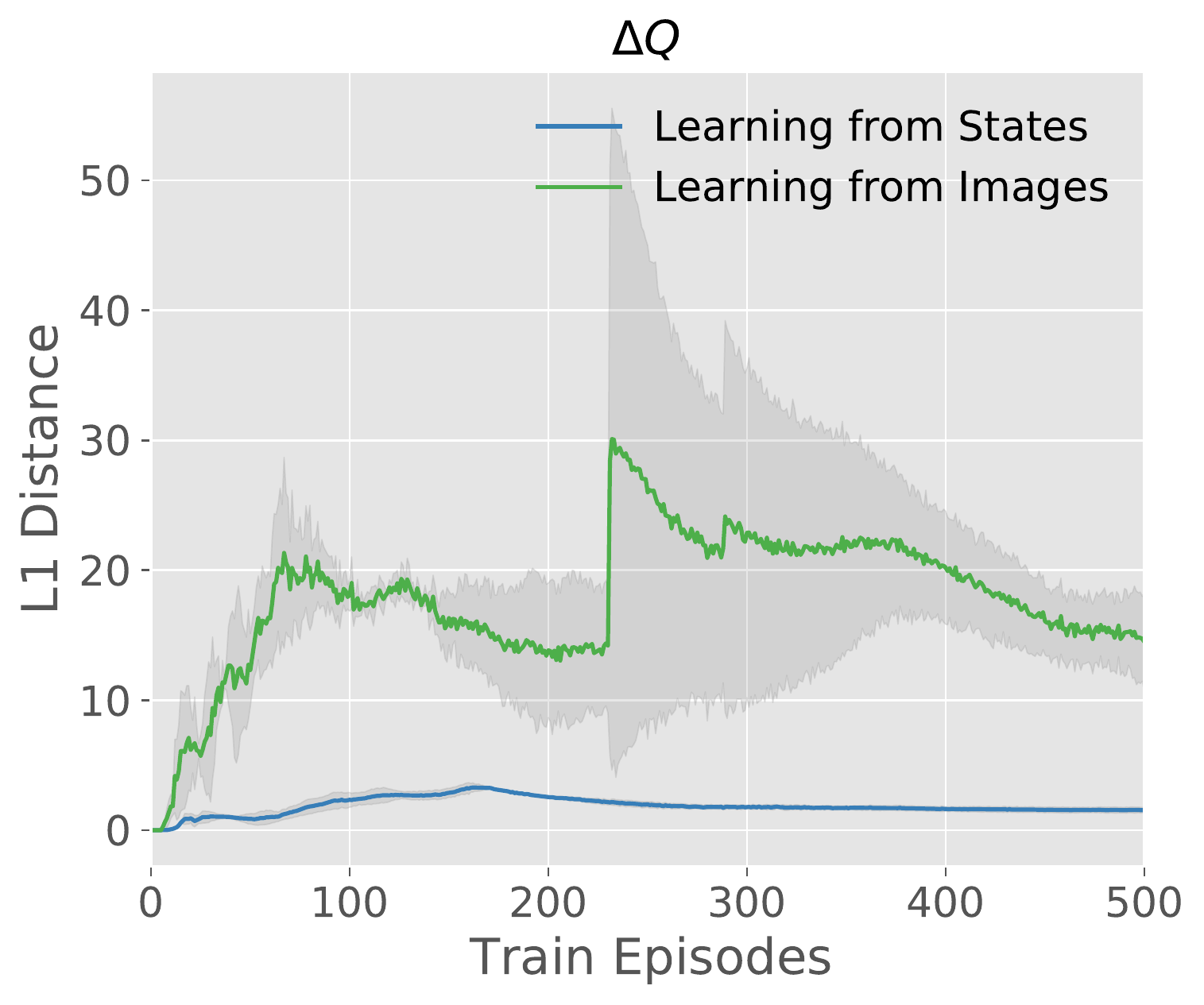}
    \caption{Difference between the Q value prediction from learned models. We trained 3 pairs of SAC agents, each pair with a same random seed, and report absolute value of difference between each pair's predicted Q values averaged on $2,000$ states encountered during training.}
    \label{fig:q_diff}
\end{figure}

We now investigate how differences in precision influence the learned neural network models. In \autoref{fig:weight_diff}, we show the L1 distance between the model weights learned with different precision -- full fp32 precision and half fp16 precision. In \autoref{fig:q_diff}, we show their difference in terms of the predicted Q value on the same state. Models trained with different precision differ in weights, and the difference grows with training. The Q value difference increases in the beginning but will eventually converge, although not to 0. The convolutional model (trained on images) has a larger difference as well as variance compared with the linear model (trained on states) in terms of both model weight and predicted Q values. 

